\documentclass[10pt,twocolumn,letterpaper]{article}

\usepackage[pagenumbers]{iccv} 

\newcommand{\RR}{\mathbb{R}}

\newcommand{\PP}{\mathbb{P}}

\definecolor{iccvblue}{rgb}{0.21,0.49,0.74}
\usepackage[pagebackref,breaklinks,colorlinks,allcolors=iccvblue]{hyperref}
\usepackage{tikz}
\usepackage{pgfplots}
\usepackage{multirow}

\def\paperID{10052} 
\def\confName{ICCV}
\def\confYear{2025}

\title{Single-Scanline Relative Pose Estimation for Rolling Shutter Cameras}

\author{Petr Hruby\\
ETH Z\"urich\\
R\"amistrasse 101, 8006 Z\"urich\\
{\tt\small petr.hruby@inf.ethz.ch}
\and
Marc Pollefeys\\
ETH Z\"urich / Microsoft Spatial AI Lab\\
R\"amistrasse 101, 8006 Z\"urich\\
{\tt\small marc.pollefeys@inf.ethz.ch}
}

\begin{document}
\maketitle
\begin{abstract}
We propose a novel approach for estimating the relative pose between rolling shutter cameras using the intersections of line projections with a single scanline per image. This allows pose estimation without explicitly modeling camera motion. Alternatively, scanlines can be selected within a single image, enabling single-view relative pose estimation for scanlines of rolling shutter cameras. Our approach is designed as a foundational building block for rolling shutter structure-from-motion (SfM), where no motion model is required, and each scanline's pose can be computed independently.
We classify minimal solvers for this problem in both generic and specialized settings, including cases with parallel lines and known gravity direction, assuming known intrinsics and no lens distortion. Furthermore, we develop minimal solvers for the parallel-lines scenario, both with and without gravity priors, by leveraging connections between this problem and the estimation of 2D structure from 1D cameras.
Experiments on rolling shutter images from the Fastec dataset demonstrate the feasibility of our approach for initializing rolling shutter SfM, highlighting its potential for further development.
The code will be made publicly available.
\end{abstract}    
\section{Introduction}
\label{sec:intro}

Relative pose estimation is a fundamental problem in computer vision with applications in Structure from Motion~\cite{schoenberger2016sfm}, SLAM, multi-view stereo, and visual odometry. This paper focuses on estimating relative pose between rolling shutter (RS) cameras.

Rolling shutter cameras capture images row by row. If the camera is static during capture, the resulting image is equivalent to that of a pinhole camera. However, if the camera moves, the rolling shutter effect distorts the image, making standard pinhole-based methods~\cite{DBLP:journals/pami/Nister04} unreliable.

A common approach to handle rolling shutter distortion is modeling camera motion with parametric models such as SLERP~\cite{DBLP:conf/siggraph/Shoemake85}, Cayley transformation~\cite{encyclopedia_of_mathematics}, linearized rotation~\cite{DBLP:conf/cvpr/AlblKP15}, affine motion~\cite{DBLP:journals/spl/SunLS16}, pushbroom~\cite{DBLP:journals/ijcv/MicusikE21}, accelerated motion~\cite{DBLP:journals/cviu/FanDW21}, or order-one motion~\cite{DBLP:journals/corr/abs-2403-11295}.  
This has been successfully applied to absolute pose estimation~\cite{DBLP:conf/eccv/Ait-AiderALM06,DBLP:conf/iros/SaurerPL15,DBLP:conf/eccv/MagerandBAP12,DBLP:journals/pami/AlblKLP20,DBLP:conf/cvpr/AlblKP16,DBLP:conf/cvpr/BaiSB22,DBLP:conf/eccv/LaoAB18,DBLP:conf/eccv/KukelovaASSP20,DBLP:conf/accv/KukelovaASP18,DBLP:conf/cvpr/Ait-AiderBA07,DBLP:conf/cvpr/AlblKP15,DBLP:journals/corr/abs-cs-0503076}, bundle adjustment~\cite{DBLP:conf/cvpr/HedborgFFR12,DBLP:conf/bmvc/NguyenL16,DBLP:conf/cvpr/LiaoQXZL23,DBLP:conf/cvpr/SaurerPL16,DBLP:conf/iclr/LiWZLL24,DBLP:conf/eccv/ZhangLXLLL24,DBLP:conf/eccv/XuLGLL24,DBLP:conf/eccv/SchubertDUSC18}, SfM~\cite{DBLP:conf/iccvw/HedborgRFF11,DBLP:conf/iccv/SaurerKBP13,DBLP:conf/ismar/KleinM09}, and motion rectification from single cameras~\cite{DBLP:conf/cvpr/RengarajanRA16,DBLP:conf/iccv/PurkaitZL17,DBLP:conf/wacv/PurkaitZ18,DBLP:journals/prl/LaoAA18,DBLP:conf/iccv/YanTZL23,DBLP:conf/cvpr/ZhuangTJCC19}, camera rigs~\cite{DBLP:conf/cvpr/AlblKLPPS20,DBLP:journals/ivc/FanDZW22,DBLP:conf/iccv/ShangRFWLZ23,DBLP:conf/iccv/Ait-AiderB09}, and videos~\cite{DBLP:conf/eccv/ZhuangT20,DBLP:conf/mmsp/JiaE12,DBLP:journals/ijcv/RingabyF12,DBLP:conf/cvpr/ForssenR10,DBLP:conf/iccv/QuLWWZ023,DBLP:journals/pami/QuLZAL23,DBLP:conf/cvpr/VasuRR18,DBLP:journals/cviu/FanDW21,DBLP:journals/spl/SunLS16,DBLP:journals/pami/FanDL23,DBLP:journals/corr/abs-2204-13886}. 

\begin{figure}
    \centering
        \includegraphics[width=0.9\linewidth]{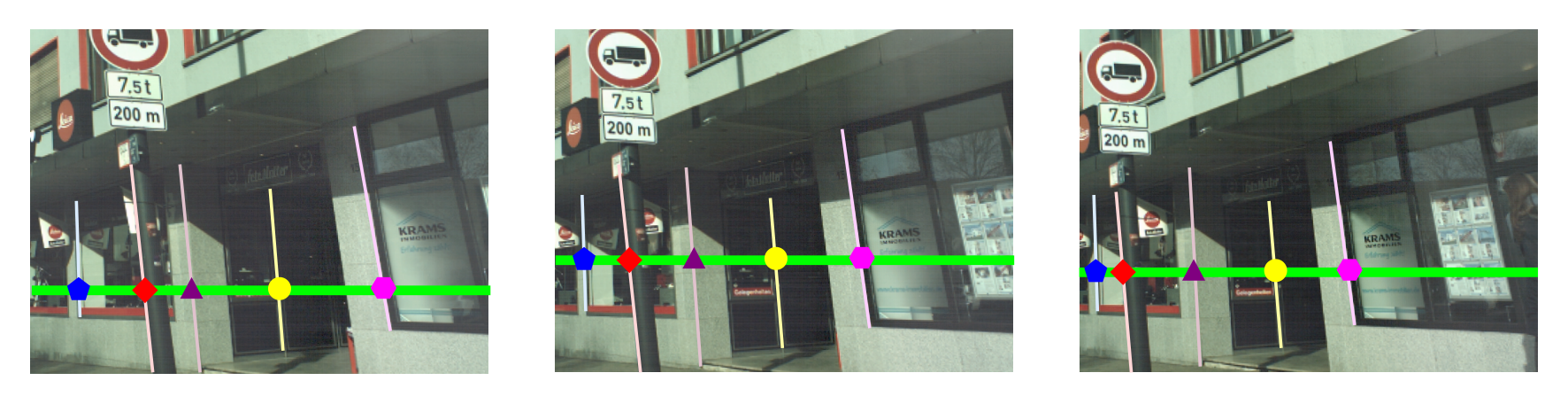}\\
        \vspace{0.07cm}
        \includegraphics[width=0.6\linewidth]{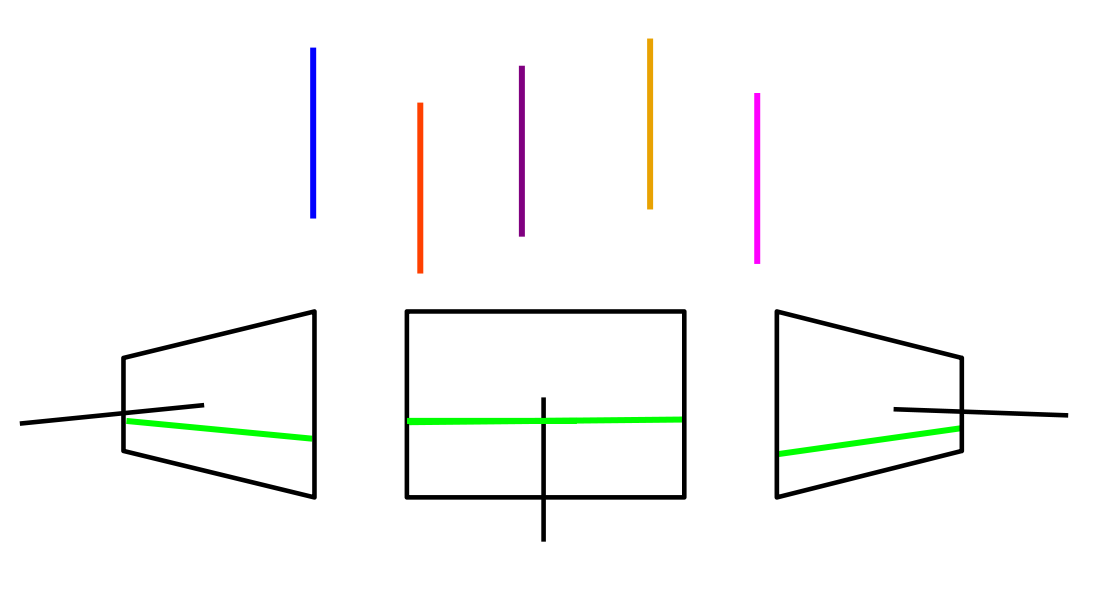}
        \vspace{-0.3cm}
    \caption{We estimate a relative pose from the intersections of the line projections with a single scanline per image (\textit{green}). This allows us to estimate relative pose between rolling shutter cameras without modelling the camera motion. }
    \label{fig:titular_image}
\end{figure}

While most methods use point-based features, some employ lines for absolute pose estimation~\cite{DBLP:conf/cvpr/Ait-AiderBA07}, bundle adjustment~\cite{DBLP:conf/eccv/ZhangLXLLL24}, and motion rectification~\cite{DBLP:conf/cvpr/RengarajanRA16,DBLP:conf/iccv/PurkaitZL17,DBLP:conf/wacv/PurkaitZ18,DBLP:journals/prl/LaoAA18}.

The problem of estimating relative pose for rolling shutter cameras has been explored in prior work. Special essential matrices for RS cameras under different motion models were introduced in~\cite{DBLP:conf/cvpr/DaiLK16}, along with linear non-minimal solvers. The problem was also solved for pure translation models with constant velocity and acceleration~\cite{DBLP:conf/iccv/ZhuangCL17}.  
IMU-based approaches reduced the required number of point correspondences from 11 to 5~\cite{DBLP:journals/corr/abs-1712-00184,DBLP:journals/corr/abs-1904-06770}. Other methods first estimate pose using a global shutter solver before refining with an RS model~\cite{DBLP:journals/corr/LeeY17a}. Additional works estimate relative pose for RS camera rigs~\cite{DBLP:conf/icip/WangFD20} and for planar scenes~\cite{9020067}. Recently,~\cite{DBLP:journals/corr/abs-2403-11295} classified RS relative pose minimal problems under the order-one motion model.  
Despite this, the general RS relative pose problem remains unsolved.

One drawback of parametric motion models is their reliance on an assumed motion that may not match real-world trajectories. An alternative is estimating pose independently for each scanline, similar to line-camera calibration~\cite{DBLP:journals/trob/HoraudML93,LI2014119,DBLP:journals/tim/LunaMLV10,DBLP:journals/sensors/SuBS18,DBLP:journals/jei/YaoZX14}. This approach has been used for tracking a single camera~\cite{DBLP:journals/tvcg/BapatDF16,DBLP:conf/cvpr/BapatPF18} and a camera rig~\cite{DBLP:journals/tvcg/DibeneMTD22} with respect to a predefined marker.

We explore the feasibility of estimating relative pose between scanlines of RS cameras, without explicit motion modelling. The goal is to use this method as a building block to initialize rolling-shutter SfM where no motion model is required and each scanline can have its pose independently computed. To the best of our knowledge, this is the first work that enables this. After the initialization, scanlines from remaining images may be registered by solving the generalized relative pose problem \cite{gen_relpose}, similarly to \cite{DBLP:journals/tvcg/BapatDF16,DBLP:conf/cvpr/BapatPF18,DBLP:journals/tvcg/DibeneMTD22}, or, in the case of a planar scene, using \cite{DBLP:conf/cvpr/BaiSB22}. Our contributions include:
\begin{itemize}
    \item Formulation of the problem of relative pose estimation from line projections in a single scanline per image, assuming known intrinsics and no lens distortion.
    \item Enumeration of minimal problems for general and specialized settings with parallel lines and gravity priors.
    \item Demonstration that the problems with parallel lines and with vertical lines and gravity prior correspond to estimating relative pose of uncalibrated cameras~\cite{DBLP:journals/pami/QuanK97}, respecively calibrated \cite{DBLP:journals/jmiv/AstromO00} 1D camera pose.
    \item Development of minimal solvers for the following cases:
    \begin{itemize}
        \item 3 cameras, 7 parallel lines (with projective ambiguity).
        \item 3 cameras, gravity prior, 7 parallel lines.
        \item 3 cameras, gravity prior, 5 vertical lines.
        \item 4 cameras, gravity prior, 4 vertical lines.
    \end{itemize}
    \item Evaluation of solvers on synthetic and real RS datasets (Fastec,~\cite{Liu2020CVPR}), taking scanlines either from \textit{different images} or from a \textit{single image}.
\end{itemize}

\subsection{Notation}
We use bold letters for matrices and vectors, and normal font for scalars. Indexing starts from one, with indices in brackets; e.g., the first element of $\mathbf{L}_{d}$ is $\mathbf{L}_{d(1)}$. Matrix indexing follows Matlab style, where $\mathbf{A}_{(:,1:2)}$ selects the first two rows of $\mathbf{A}$. The scanlines are denoted with a scalar $y \in \RR$ giving their y-coordinate. The set of all 3D lines is denoted as $Gr(1,3)$.

\section{Problem Formulation}
\label{sec:formulation}

This section describes the geometry of a line projected into a rolling-shutter camera and derives an algebraic constraint useful for estimating camera poses.

Consider $n$ rolling-shutter cameras capturing the same 3D scene. Due to motion. the cameras capture different rows at different poses. Let $\mathbf{R}_i(y) \in SO(3)$ and $\mathbf{C}_i(y) \in \mathbb{R}^3$ denote the orientation and center of camera $i$ at row $y$.

We aim to estimate the relative pose between the cameras without explicitly modeling their motion functions. To achieve this, we select a single row $y_i$ per camera and measure intersections between these selected rows and the projections of $m$ straight lines $\mathbf{L}_j \in Gr(1,3)$. Then, we use this information to recover the camera poses $\mathbf{R}_i(y_i)$, $\mathbf{C}_i(y_i)$. Figure~\ref{fig:titular_image} illustrates the setup.

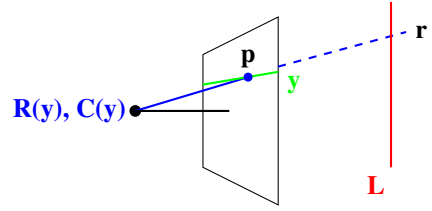
\begin{figure}
    \centering
    \begin{tikzpicture}

    \coordinate (C) at (-1.4,0.25);
    \coordinate (p) at (0.1,0.7);
    \coordinate (r1) at (-2,-0.2);
    \coordinate (r2) at (2.2,1.3);
    \coordinate (y1) at (-0.5,0.6);
    \coordinate (y2) at (0.5,0.77);
    
    \filldraw[black] (C) circle (2pt);
    \node[left, blue] at (C) {\textbf{R(y), C(y)}};
    
    \draw[black] (-0.5,-0.5) -- (-0.5,1) -- (0.5,1.5) -- (0.5,-1) -- cycle;

    \draw[blue, thick] (C) -- (p);
    \draw[blue, dashed, thick] (0.5,0.84) -- (r2);
    \node[right] at (r2) {\textbf{r}};
    
    \draw[green, thick] (y1) -- (y2);
    \node[right, green] at (0.5,0.6) {\textbf{y}};

    \filldraw[blue] (p) circle (1.5pt);
    \node[above] at (p) {\textbf{p}};

    \draw[black, thick] (C) -- (-0.15,0.25);
    
    \draw[red, thick] (2.0,-0.5) -- (2.0,1.7);
    \node[below, red] at (1.8,-0.5) {\textbf{L}};

\end{tikzpicture}
    \caption{Illustration of the line projection. Line $\mathbf{L}$ projects onto scanline $\mathbf{y}$ at point $\mathbf{p}$. Then, the ray $\mathbf{r}$ obtained by backprojecting $\mathbf{p}$ intersects $\mathbf{L}$, yielding the constraint \eqref{eq:the_constraint}. }
    \label{fig:projection_constraint}
\end{figure}

Let $\mathbf{p}_{i,j} \in \mathbb{P}^2$, with $p_{i,j} = [x_{i,j} \ y_i \ 1]^T$, denote the projection of line $\mathbf{L}_j$ onto row $y_i$ of camera $i$. Then, the back-projected ray $\mathbf{r}_{i,j}$ from $\mathbf{p}_{i,j}$ intersects line $\mathbf{L}_j$, giving:
\begin{equation}
    \mathbf{r}_{i,j} = \mathbf{C}_i(y_i) + \lambda_{i,j} \mathbf{R}_i(y_i)^T \mathbf{p}_{i,j}, \quad \lambda_{i,j} \in \mathbb{R}.
\end{equation}
This situation is depicted in Fig.~\ref{fig:projection_constraint}.
Let us represent $\mathbf{L}_j$ with a pair $(\mathbf{L}_{0,j}, \mathbf{L}_{d,j})$, where $\mathbf{L}_{0,j}$ is a point on the line and $\mathbf{L}_{d,j}$ is its direction. The ray $\mathbf{r}_{i,j}$ intersects $\mathbf{L}_j$ if and only if:
\begin{equation}
    \exists \alpha, \beta \in \mathbb{R}: \quad \mathbf{C}_i(y_i) + \alpha \mathbf{R}_i(y_i)^T \mathbf{p}_{i,j} = \mathbf{L}_{0,j} + \beta \mathbf{L}_{d,j}.
\end{equation}
This is equivalent to requiring that vectors $\mathbf{R}_i(y_i)^T \mathbf{p}_{i,j}$, $\mathbf{L}_{d,j}$, and $\mathbf{L}_{0,j} - \mathbf{C}_i(y_i)$ are linearly dependent. Since three 3D vectors are linearly dependent if and only if their scalar triple product vanishes, we obtain:
\begin{equation}
    \mathbf{p}_{i,j}^T \mathbf{R}_i(y_i) [\mathbf{L}_{d,j}]_{\times} (\mathbf{L}_{0,j} - \mathbf{C}_i(y_i)) = 0. \label{eq:the_constraint}
\end{equation}

%
Each projection provides a single constraint on the unknowns (lines and camera poses). A 3D line has four degrees of freedom, requiring four scanlines for triangulation and at least five scanlines for constraining relative pose.

\subsection{Assuming parallel lines}

Let us consider the scenario where all lines are parallel, which frequently occurs in real-world settings. Here, all lines share a common direction $\mathbf{L}_d \in \mathbb{R}^3$. Since Eq.~\eqref{eq:the_constraint} is invariant to scene rotation, we align $\mathbf{L}_d$ with the y-axis, $\mathbf{e}_2$. This allows representing each line $\mathbf{L}_j$ with a single point $\mathbf{L}_{0,j}$ in the xz-plane, with its y-coordinate equal to zero.

With a shared direction, each line has only two individual degrees of freedom (DoF). Once $\mathbf{L}_d$ is known, a line can be triangulated from just two cameras. This reduces the minimum number of cameras needed for pose estimation to three.
To simplify Eq.~\eqref{eq:the_constraint}, we substitute $\mathbf{e}_2$ for $\mathbf{L}_{d,j}$:
\begin{equation}
    \mathbf{p}_{i,j}^T \mathbf{R}_i(y_i) \left( \begin{bmatrix}  \mathbf{L}_{0,j(3)} \\ 0 \\ -\mathbf{L}_{0,j(1)} \end{bmatrix} - \begin{bmatrix}  \mathbf{C}_i(y_i)_{(3)} \\ 0 \\ -\mathbf{C}_i(y_i)_{(1)} \end{bmatrix} \right) = 0. \label{eq:matrix_form_00}
\end{equation}
We define rotation $\mathbf{R}_{0,i} \in SO(3)$ around the x-axis, which maps row $y_i$ to $y=0$, and set $\mathbf{p}'_{i,j} = \mathbf{R}_{0,i} \mathbf{p}_{i,j}$. Then, there holds $\mathbf{p}'_{i,j} = [x_{i,j}' \ 0 \ 1]^T$.
Defining
\begin{equation}
\begin{split}
    \mathbf{R}_i' = \mathbf{R}_{0,i} \mathbf{R}_i(y_i),\\ \quad \mathbf{t}_i' = -\mathbf{R}_{0,i} \mathbf{R}_i(y_i) [ \mathbf{C}_i(y_i)_{(3)} \ 0 \ -\mathbf{C}_i(y_i)_{(1)}]^T,
\end{split}
\label{eq:R_prime_definition}
\end{equation}
and substituting $\mathbf{R}_i'$, $\mathbf{t}_i'$, and $\mathbf{p}'_{i,j}$ into Eq.~\eqref{eq:matrix_form_00}, we obtain:
\begin{equation}
    \begin{bmatrix} x_{i,j}' & 1 \end{bmatrix}^T \begin{bmatrix} -\mathbf{R}'_{i(1,3)} & \mathbf{R}'_{i(1,1)} &  \mathbf{t}'_{i(0)} \\ -\mathbf{R}'_{i(3,3)} & \mathbf{R}'_{i(3,1)} &  \mathbf{t}'_{i(2)} \end{bmatrix} \begin{bmatrix}  \mathbf{L}_{0,j(1)} \\ \mathbf{L}_{0,j(3)} \\ 1 \end{bmatrix} = 0. \label{eq:matrix_form_03}
\end{equation}
We can write Equation \eqref{eq:matrix_form_03} in a vector form as
\begin{equation}
    \mathbf{u}_{i,j}^T \mathbf{A}_{i} \mathbf{L}_{h,j} = 0, \label{eq:matrix_form_04}
\end{equation}
where \(\mathbf{A}_i \in \mathbb{R}^{2,3}\) is a matrix encoding the camera pose, \(\mathbf{u}_{i,j} \in \mathbb{P}^1\) is a vector encoding the observation, and \(\mathbf{L}_{h,j} \in \mathbb{P}^2\) is a homogeneous representation of point \(\mathbf{L}_{0,j}\). Alternatively, we define \(\mathbf{u}'_{i,j} = [1 \ -x_{i,j}']^T\), and rewrite \eqref{eq:matrix_form_04} as
\begin{equation}
    \lambda_{i,j} \mathbf{u}'_{i,j} = \mathbf{A}_{i} \mathbf{L}_{j,h}, \quad \lambda_{i,j} \in \mathbb{R}. \label{eq:matrix_form_05}
\end{equation}
Since constraint \eqref{eq:matrix_form_04} is homogeneous, matrix \(\mathbf{A}_i\) has \(5\) DoF. The elements \(\mathbf{R}_i(y_i)\), \(\mathbf{C}_i(y_i)_{(1)}\), \(\mathbf{C}_i(y_i)_{(3)}\) of the pose, which appear in \(\mathbf{A}_i\), also have \(5\) DoF. Using the Gr\"obner basis method, we have verified that for every matrix \(\mathbf{A} \in \mathbb{R}^{2,3}\), there are \(8\) ways to decompose it into pose parameters \(R \in SO(3)\), \(C_{(1)} \in \mathbb{R}\), \(C_{(3)} \in \mathbb{R}\), and a scale factor \(\sigma \in \mathbb{R}\). This means that \(\mathbf{A}_i\) has no internal constraints, and every \(2\times3\) matrix can represent \(\mathbf{A}_i\) for some pose. 

Therefore, the relative pose problem can be transformed into finding matrices \(\mathbf{A}_i \in \mathbb{R}^{2,3}\), vectors \(\mathbf{L}_{h,j} \in \mathbb{P}^2\), and scalars \(\lambda_{i,j} \in \mathbb{R}\) such that \eqref{eq:matrix_form_05} holds for every \(i \in \{1,...,n\}\), \(j \in \{1,...,m\}\). This is equivalent to the problem of estimating 2D structure from images taken by \(n\) uncalibrated 1D cameras, which has been studied in \cite{DBLP:journals/pami/QuanK97}.

\subsubsection{Ambiguities in the case with parallel lines}

The assumption that all lines are parallel introduces ambiguities into the relative pose estimation. First, since all line directions \(\mathbf{L}_{d,j}\) are equal, the values of constraints \eqref{eq:the_constraint} remain unchanged if we shift \(\mathbf{C}_i(y_i)\) to \(\mathbf{C}_i(y_i) + \gamma \mathbf{L}_{j,d}\) for any \(\gamma \in \mathbb{R}\). Therefore, it is only possible to estimate the camera centers up to translation along \(\mathbf{L}_d\).

Furthermore, if \eqref{eq:matrix_form_05} holds for some \(\mathbf{A}_i \in \mathbb{R}^{2,3}\), \(\mathbf{L}_{h,j} \in \mathbb{P}^2\), then \eqref{eq:matrix_form_05} also holds for \(\mathbf{A}_i \cdot \mathbf{H}^{-1}\), \(\mathbf{H} \cdot \mathbf{L}_{j,h}\) for every invertible \(\mathbf{H} \in \mathbb{R}^{3,3}\). Therefore, the cameras and the structure can only be estimated up to a projective transformation \(\mathbf{H}\).

We show later in the paper that we can resolve the projective ambiguity by introducing priors on the cameras. To resolve the ambiguity in the camera centers, we can, for instance, observe one line orthogonal to \(\mathbf{L}_d\) and use it to estimate the camera centers.

\section{Enumerating minimal problems} \label{sec:enumerating}
In this section, we classify minimal problems for estimating the relative pose from projections of $n$ lines onto $m$ scanlines, along with their degrees. We consider five settings:

\begin{enumerate}[label=\Alph*.]
    \item \textbf{Fully generic setting}
    \item \textbf{Parallel lines}
    \item \textbf{Gravity prior for cameras + Generic lines}
    \item \textbf{Gravity prior for cameras + Parallel lines}
    \item \textbf{Gravity prior for cameras + Vertical lines}
\end{enumerate}

In cases A and C, we estimate the pose up to global rotation, translation, and scale. In cases D and E, we estimate the pose up to these transformations and an additional shift of centers $\mathbf{C}_i(y_i)$ along direction $\mathbf{L}_d$. In case B, we estimate the pose up to these transformations, the shift of centers along $\mathbf{L}_d$, and the projective transformation $\mathbf{H}$.

For each setting, we determine the number of cameras $m$ and the number of lines $n$ s.t. the problem \eqref{eq:the_constraint} is minimal. We assume \textit{complete visibility}, meaning each line is projected into each camera. We follow concepts from \cite{DBLP:journals/pami/DuffKLP24}:

\noindent\textit{Minimal problem} \cite[Def.~1]{DBLP:journals/pami/DuffKLP24}: A parametric polynomial system $f(\theta,\mathbf{x})=0$ with a finite nonzero number of solutions $\mathbf{x}^*$ for a generic set of parameters $\theta$.

\noindent\textit{Balanced problem} \cite[Def.~2]{DBLP:journals/pami/DuffKLP24}: A parametric polynomial system $f(\theta,\mathbf{x})=0$ where the dimensionality of the space of variables $\mathbf{x}$ equals the number of independent equations.

\noindent\textit{Condition for minimal problems} \cite[Step.~1, Corr.~2]{DBLP:journals/pami/DuffKLP24}: A problem $f(\theta,\mathbf{x})=0$ is minimal if and only if it is \textit{balanced} and its Jacobian $\frac{\partial f(\theta, \mathbf{x}^*)}{\partial \mathbf{x}}$ is invertible for generic $\theta$, $\mathbf{x}^*$.

This summary provides the foundation for enumerating minimal problems. For detailed theory, refer to \cite{DBLP:journals/pami/DuffKLP24}. In our case, the parameter vector $\theta$ consists of projections $\mathbf{p}_{i,j} \in \PP^2$, and the variable vector $\mathbf{x}$ encodes camera poses $(\mathbf{R}_i(y_i), \mathbf{C}_i(y_i))$ and lines $(\mathbf{L}_{0,j}, \mathbf{L}_{d,j})$. The polynomial system $f(\mathbf{x},\theta)$ consists of equations in the form of \eqref{eq:the_constraint}. Following \cite{DBLP:journals/pami/DuffKLP24}, we used the following steps to list all minimal problems and determine their degrees:

\begin{itemize}
    \item Enumerate all balanced problems for each setting A–E.
    \item For each balanced problem, generate random camera poses and lines $\mathbf{x}^*$, reproject the lines into the cameras, and obtain projections $\theta$ such that $f(\theta, \mathbf{x}^*) = 0$.
    \item Compute the Jacobian $\frac{\partial f(\theta, \mathbf{x}^*)}{\partial \mathbf{x}}$. Label the problem as minimal if $\det \frac{\partial f(\theta, \mathbf{x}^*)}{\partial \mathbf{x}} \neq 0$.
    \item Use monodromy computation from \cite{DBLP:conf/icms/BreidingT18} to determine the degree of minimal problems.
\end{itemize}
According to algebraic geometry, finding a single instance where the Jacobian is full-rank suffices to claim generic minimality.
%
In the following subsections, we enumerate minimal problems for each setting A–E.

\begin{table}
    \centering
    \renewcommand{\arraystretch}{0.95} 
    \setlength{\tabcolsep}{5pt} 
    \begin{tabular}{c c c c c ||  c c c c c}
        \toprule
        \midrule
        S. & $m$ & $n$ & Min. & Deg & S. & $m$ & $n$ & Min. & Deg \\
        \midrule
        A & $5$ & $23$ & Y & 389k+ & D & $3$ & $7$ & Y & $48$ \\
        A & $21$ & $7$ & Y & 40k+  & D & $4$ & $5$ & Y & $232$ \\
        \cline{1-5}
        B & $3$ & $7$ & Y & $2^*$  & D & $6$ & $4$ & Y & $1224$ \\
        \cline{6-10}
        B & $4$ & $6$ & Y & $2^*$  & E & $3$ & $5$ & Y & $16$ \\
        \cline{1-5}
        C & $5$ & $15$ & Y & 1.8M+  & E & $4$ & $4$ & Y & $32$ \\
        C & $15$ & $5$ & Y & 532k+  &  &  &  &  &  \\
        \bottomrule
    \end{tabular}
    \vspace{-0.3cm}
    \caption{List of balanced problems for settings A–E. \textit{S.} denotes the setting, $m$ and $n$ are the number of cameras and lines, and Deg is the degree. Column Min. contains Y if the problem is minimal. $2^*$ indicates two solutions in terms of $2 \times 3$ matrices, and $+$ denotes computations that did not terminate. See Sec.~\ref{sec:enumerating} for details. }
    \label{tab:balanced_problems}
\end{table}

\paragraph{A. Fully Generic Setting} Each camera has \( 6 \) degrees of freedom (DoF), while each line has \( 4 \) DoF. Due to gauge freedom, we fix \( 7 \) DoF (\( 3 \) for rotation, \( 3 \) for translation, and \( 1 \) for scale). There are \( m \cdot n \) constraints in the form of \eqref{eq:the_constraint}, fixing \( m \cdot n \) DoF. The balanced constraint is therefore:
\begin{equation}
    6m + 4n - 7 = mn. \label{eq:balanced_A}
\end{equation}
We define a function \( n_A(m) \), mapping the number of cameras \( m \) to the number of lines \( n \), s.t. \eqref{eq:balanced_A} holds for $m$, $n_A(m)$:
\begin{equation}
    n_A(m) = \frac{6m - 7}{m - 4}.
\end{equation}
As stated in Sec.~\ref{sec:formulation}, the minimal number of cameras is \( m \geq 5 \). The function \( n_A(m) \) is strictly decreasing for \( m > 4 \), and since \( \lim_{m \to \infty} n_A(m) = 6 \), no balanced problem has fewer than \( 7 \) lines. Evaluating \( n_A(m) \) for integer values of \( m \) from \( 5 \) onward, and searching for integer values of \( n_A(m) \), we identify the following balanced cases:

\begin{itemize}
    \item \( m = 5, n = 23; \quad m = 21, n = 7 \)
\end{itemize}

\paragraph{B. Parallel Lines} As discussed in Sec.~\ref{sec:formulation}, the pose is estimated up to a shift along \( \mathbf{L}_d \) and a projective transformation \( \mathbf{H} \in \mathbb{R}^{3 \times 3} \). The pose estimation corresponds to estimating matrices \( \mathbf{A}_i \in \mathbb{R}^{2 \times 3} \) and points \( \mathbf{L}_{h,j} \in \mathbb{P}^2 \) such that \eqref{eq:matrix_form_04} holds for all \( i, j \). Here, each camera has \( 5 \) DoF, each line has \( 2 \) DoF, and we fix \( 8 \) DoF due to projective ambiguity. The balanced constraint is:
\begin{equation}
    5m + 2n - 8 = mn. \label{eq:balanced_B}
\end{equation}
Defining \( n_B(m) \) as:
\begin{equation}
    n_B(m) = \frac{5m - 8}{m - 2},
\end{equation}
and noting that \( m \geq 3 \), we find \( \lim_{m \to \infty} n_B(m) = 5 \), implying no balanced problem has fewer than \( 6 \) lines. Evaluating \( n_B(m) \) for integer \( m \), we obtain:

\begin{itemize}
    \item \( m = 3, n = 7; \quad  m = 4, n = 6 \)
\end{itemize}

\paragraph{C. Gravity Prior for Cameras + Generic Lines} With a known vertical direction, each camera has \( 4 \) DoF (\( 1 \) for rotation, \( 3 \) for translation), and we fix \( 5 \) DoF due to gauge freedom. Lines remain generic, contributing \( 4 \) DoF each. The balanced constraint is:

\begin{equation}
    4m + 4n - 5 = mn. \label{eq:balanced_C}
\end{equation}
The corresponding function is:
\begin{equation}
    n_C(m) = \frac{4m - 5}{m - 4}.
\end{equation}
We find \( \lim_{m \to \infty} n_C(m) = 4 \), so no balanced problem has fewer than \( 5 \) lines. Checking integer values of $m$, we get:
\begin{itemize}
    \item \( m = 5, n = 15; \quad  m = 15, n = 5 \)
\end{itemize}

\paragraph{D. Gravity Prior for Cameras + Parallel Lines} Here, camera rotation has \( 1 \) DoF, and the camera center has \( 2 \) DoF (up to translation along the line). We fix \( 4 \) DoF due to gauge freedom. Lines share \( 2 \) DoF for direction \( \mathbf{L}_d \), with each line having \( 2 \) additional DoF. The balanced constraint is:
\begin{equation}
    3m + 2 + 2n - 4 = mn.
\end{equation}
We get \( n_D(m) \) as:
\begin{equation}
    n_D(m) = \frac{3m - 2}{m - 2}.
\end{equation}
For \( m \geq 3 \), we find \( \lim_{m \to \infty} n_D(m) = 3 \), meaning no balanced problem has fewer than \( 4 \) lines. We have found:

\begin{itemize}
    \item \( m = 3, n = 7; \quad m = 4, n = 5; \quad m = 6, n = 4 \)
\end{itemize}

\paragraph{E. Gravity Prior for Cameras + Vertical Lines} Similar to the previous case, each camera has \( 3 \) DoF, and we fix \( 4 \) DoF due to gauge freedom. Since the direction \( \mathbf{L}_d \) is fixed, each line has \( 2 \) DoF. The balanced constraint is:

\begin{equation}
    3m + 2n - 4 = mn.
\end{equation}
We define \( n_E(m) \) as:
\begin{equation}
    n_E(m) = \frac{3m - 4}{m - 2}.
\end{equation}
For \( m \geq 3 \), we find \( \lim_{m \to \infty} n_E(m) = 3 \), meaning no balanced problem has fewer than \( 4 \) lines. We found problems:

\begin{itemize}
    \item \( m = 3, n = 5; \quad m = 4, n = 4 \)
\end{itemize}

All balanced problems for these settings are summarized in Table~\ref{tab:balanced_problems}, along with their minimality status and degree. The table shows that problems with parallel/vertical lines (B, D, E) are minimal and computationally feasible. Problems with generic lines (A, C) are minimal but have degrees too high for practical RANSAC solutions. For problems $(A,5,23)$ and $(C,5,15)$, monodromy computations did not terminate within a month.

\section{Minimal solvers}

In this section, we introduce solvers for the minimal problems $(B,3,7)$, $(D,3,7)$, $(E,3,5)$, $(E,4,4)$ from Section~\ref{sec:enumerating}. We use the notation and concepts from Sec.~\ref{sec:formulation}.

\subsection{Parallel lines with generic cameras} \label{sec:parallel_lines}

Here, we describe the solution to the problem $(B,3,7)$. Since the lines are parallel, our goal is to find matrices $\mathbf{A}_1, \mathbf{A}_2, \mathbf{A}_3 \in \mathbb{R}^{2 \times 3}$ and vectors $\mathbf{L}_{h,j}$ for $j \in \{1, \dots, 7\}$, such that equation \eqref{eq:matrix_form_04} holds for every camera $i$ and line $j$. Due to projective ambiguity, we aim to recover the camera poses up to a common projective transformation $\mathbf{H} \in \mathbb{R}^{3 \times 3}$. Each camera triplet $\mathbf{A}_1, \mathbf{A}_2, \mathbf{A}_3$ can be expressed as:
\begin{equation}
    \mathbf{A}_1 = \mathbf{A}_{c,1} \cdot \mathbf{H}, \quad 
    \mathbf{A}_2 = \mathbf{A}_{c,2} \cdot \mathbf{H}, \quad 
    \mathbf{A}_3 = \mathbf{A}_{c,3} \cdot \mathbf{H},
\end{equation}
for some $\mathbf{H} \in \mathbb{R}^{3 \times 3}$, with $\mathbf{A}_{c,1}, \mathbf{A}_{c,2}, \mathbf{A}_{c,3}$ taking the form:
\begin{equation}
    \mathbf{A}_{c,1} = 
    \left[\begin{smallmatrix}
        1 & 0 & 0 \\ 
        \alpha_1 & \alpha_1 & \alpha_1
    \end{smallmatrix}\right],
    \mathbf{A}_{c,2} = 
    \left[\begin{smallmatrix}
        1 & 0 & 0 \\ 
        \alpha_2 & \alpha_3 & \alpha_4
    \end{smallmatrix}\right],
    \mathbf{A}_{c,3} = 
    \left[\begin{smallmatrix}
        1 & 0 & 0 \\ 
        \alpha_5 & \alpha_6 & \alpha_7
    \end{smallmatrix}\right]. \label{eq:canonical_cameras}
\end{equation}
The proof follows analogously to Proposition~2 in \cite{DBLP:conf/cvpr/HrubyKDOPPL23}. These canonical-form matrices have $7$ DoF.

To estimate $\mathbf{A}_{c,1}, \mathbf{A}_{c,2}, \mathbf{A}_{c,3}$, we leverage the fact that, as shown in \cite{DBLP:journals/pami/QuanK97}, a trifocal tensor exists between three camera matrices $\mathbf{A}_i \in \mathbb{R}^{2 \times 3}$. To compute this tensor, we stack the constraints from \eqref{eq:matrix_form_05} into the matrix equation:
\begin{equation}
    \begin{bmatrix}
        \mathbf{A}_1 & \mathbf{u}'_{1,j} & 0 & 0 \\
        \mathbf{A}_2 & 0 & \mathbf{u}'_{2,j} & 0 \\
        \mathbf{A}_3 & 0 & 0 & \mathbf{u}'_{3,j} 
    \end{bmatrix}
    \begin{bmatrix}
        \mathbf{L}_{h,j} \\ -\lambda_{1,j} \\ -\lambda_{2,j} \\ -\lambda_{3,j}
    \end{bmatrix} = 0. \label{eq:det_00}
\end{equation}
For this equation to hold, the determinant of the coefficient matrix must vanish, leading to the constraint:
\begin{equation}
    \sum_{a = 1}^{2} \sum_{b = 1}^{2} \sum_{c = 1}^{2} 
    \mathbf{u}'_{1,j(a)} \mathbf{u}'_{2,j(b)} \mathbf{u}'_{3,j(c)} 
    \mathbf{T}_{(a,b,c)} = 0, \label{eq:tensor_constraint}
\end{equation}
where $\mathbf{T} \in \mathbb{R}^{2 \times 2 \times 2}$ is the trifocal tensor with elements:
\begin{equation}
    \mathbf{T}_{(a,b,c)} = (-1)^{a+b+c} 
    \det\left( 
        \begin{bmatrix}
            \mathbf{A}_{1(a,:)}^T \ \mathbf{A}_{2(b,:)}^T \ \mathbf{A}_{3(c,:)}^T 
        \end{bmatrix} 
    \right). \label{eq:tensor_elements}
\end{equation}
As shown in \cite{DBLP:journals/pami/QuanK97}, $\mathbf{T}$ has no internal constraints and it can be estimated using $7$ line correspondences.

Since $\mathbf{T}$ is homogeneous, it has $7$ DoF, matching the DoF of the canonical camera matrices. We verified that for every tensor $\mathbf{T} \in \mathbb{R}^{2 \times 2 \times 2}$, there are exactly two corresponding triplets of matrices $\mathbf{A}_{c,1}, \mathbf{A}_{c,2}, \mathbf{A}_{c,3}$. The proof and the decomposition algorithm for recovering these matrices can be found in the SM. Once $\mathbf{A}_{c,1}, \mathbf{A}_{c,2}, \mathbf{A}_{c,3}$ are found, we can triangulate the lines up to the projective transformation $\mathbf{H}$.

The solver for the $(B,3,7)$ problem, which estimates the trifocal tensor $\mathbf{T}$, operates as follows:

\begin{enumerate}
    \item \textbf{Input:} Projections $\mathbf{p}_{i,j}$ of $7$ lines in $3$ cameras.
    \item Compute matrices $\mathbf{R}_{0,i}$ and vectors $\mathbf{u}_{i,j}$ and $\mathbf{u}'_{i,j}$.
    \item Construct the linear system \eqref{eq:tensor_constraint}.
    \item Solve \eqref{eq:tensor_constraint} using singular value decomposition (SVD) to estimate the tensor $\mathbf{T}$.
\end{enumerate}
On average, the solver runs in $6.92 \, \mu s$.

\subsection{Vertical lines and known gravity} \label{sec:vertical_lines}

Here, we analyze the solutions to the problems $(E,3,5)$ and $(E,4,4)$. We assume that that all lines are vertical and the gravity directions $\mathbf{v}_i \in \PP^2$ for the cameras are known.


Similar to Sec.~\ref{sec:parallel_lines}, we set $\mathbf{L}_{d,j} = \mathbf{e}_2$. Therefore, Equation \eqref{eq:matrix_form_03} still holds. Furthermore the vanishing points $\mathbf{v}_i$ for $i \in \{1,...,m\}$ associated with the vertical lines are known, and there holds $\mathbf{v}_i = \mathbf{R}_i(y_i) \mathbf{e}_2$. We can decompose $\mathbf{R}_i(y_i) = \mathbf{R}_{v,i} \mathbf{R}_{y,i}$, where $\mathbf{R}_{y,i}$ is a rotation around y-axis, and $\mathbf{R}_{v,i}$ is a rotation around an axis in the xz-plane, such that $\mathbf{R}_{v,i} \mathbf{e}_2 = \mathbf{v}_i$. Since $\mathbf{R}_{v,i}$ is calculated directly from $\mathbf{v}_i$, we assume it to be known. 

In this case, $\mathbf{R}'_i$ \eqref{eq:R_prime_definition} takes the form $\mathbf{R}'_i = \mathbf{R}_{0,i} \mathbf{R}_{v,i} \mathbf{R}_{y,i}$.
While both $\mathbf{R}_{0,i}$ and $\mathbf{R}_{v,i}$ have their axes in the xz-plane, $\mathbf{R}_{0,i} \mathbf{R}_{v,i}$ does generically not share this property. However, we can decompose $\mathbf{R}_{0,i} \mathbf{R}_{v,i} = \mathbf{R}_{a,i} \mathbf{R}_{b,i}$, s.t. $\mathbf{R}_{a,i}$ has its axis in the xz-plane and $\mathbf{R}_{b,i}$ rotates around the y-axis.


Now, we decompose $\mathbf{R}'_i$ as $\mathbf{R}_{a,i} \cdot \mathbf{R}''_i$, where $\mathbf{R}_{a,i}$ is a known rotation around axis in the xz-plane, and $\mathbf{R}''_i = \mathbf{R}_{b,i} \cdot \mathbf{R}_{y,i}$ is an unknown rotation around y-axis.

We represent $\mathbf{R}_{a,i}, \mathbf{R}''_i$ with quaternions
        $\mathbf{q}_{i} = [\mathbf{q}_{i,w} \ \mathbf{q}_{i,x} \ 0 \ \mathbf{q}_{i,z}]^T, \
\mathbf{q}''_{i} = [\mathbf{q}''_{i,w} \ 0 \ \mathbf{q}''_{i,y} \ 0]^T,$
write the elements of $\mathbf{R'}$ in the form
\begin{equation}
    \begin{split}
        \mathbf{R}'_{i(1,1)} = (1-2\mathbf{q}_{i,z}^2) (1-2 \mathbf{q}_{i,y}''^2) - 4 \mathbf{q}_{i,x} \mathbf{q}_{i,z} \mathbf{q}''_{i,w} \mathbf{q}_{i,y}'',\\
        \mathbf{R}'_{i(1,3)} = 2 (1-2\mathbf{q}_{i,z}^2) \mathbf{q}_{i,w}'' \mathbf{q}_{i,y}'' + 2 \mathbf{q}_{i,x} \mathbf{q}_{i,z} (1-2 \mathbf{q}_{i,y}''^2),\\
        \mathbf{R}'_{i(3,1)} = 2 \mathbf{q}_{i,x} \mathbf{q}_{i,z} (1-2 \mathbf{q}_{i,y}''^2) - 2 (1-2\mathbf{q}_{i,x}^2) \mathbf{q}_{i,w}'' \mathbf{q}_{i,y}'', \\
        \mathbf{R}'_{i(3,3)} = (1-2\mathbf{q}_{i,x}^2) (1-2 \mathbf{q}_{i,y}''^2) + 4 \mathbf{q}_{i,x} \mathbf{q}_{i,z} \mathbf{q}''_{i,w} \mathbf{q}_{i,y}'',
    \end{split}
\end{equation}
and decompose the left $2 \times 2$ submatrix of $\mathbf{A}_i$ \eqref{eq:matrix_form_04} as
\begin{equation}
    \begin{bmatrix}
        -\mathbf{R}'_{i,(1,3)} & \mathbf{R}'_{i,(1,1)} \\ -\mathbf{R}'_{i,(3,3)} & \mathbf{R}'_{i,(3,1)}
    \end{bmatrix} = 
    \mathbf{A}_{i,a} \cdot \mathbf{A}_{i,b},
\end{equation}
with
\begin{equation}
    \mathbf{A}_{i,a} = 
    \begin{bmatrix}
        -2\mathbf{q}_{i,x} \mathbf{q}_{i,z} & (1-2\mathbf{q}_{i,z}^2) \\ -(1-2\mathbf{q}_{i,x}^2) & 2\mathbf{q}_{i,x} \mathbf{q}_{i,z}
    \end{bmatrix}, \label{eq:Aia}
\end{equation}
\begin{equation}
    \mathbf{A}_{i,b} = \begin{bmatrix}
        (1-2 \mathbf{q}_{i,y}''^2) & 2 q''_{i,w} \mathbf{q}_{i,y}'' \\ -2q''_{i,w} \mathbf{q}_{i,y}'' & (1-2 \mathbf{q}_{i,y}''^2)
    \end{bmatrix}.
\end{equation}
$\mathbf{A}_{i,a}$ can be directly obtained from the known gravity direction and the current scanline $y_i$, and $\mathbf{A}_{i,b}$ is an unknown $2D$ rotation matrix. We define $\mathbf{A}_{i}' = \mathbf{A}_{i,a}^{-1} \mathbf{A}_i$, and $\mathbf{u}_{i,j}'' = \mathbf{A}'^T_{i,a} \mathbf{u}_{i,j}$. Then, there holds
\begin{equation}
    \mathbf{u}_{i,j}''^T \mathbf{A}'_{i} \mathbf{L}_{j,h} = 0. \label{eq:matrix_constraint_calibrated}
\end{equation}
Since the left $2 \times 2$ submatrix of $\mathbf{A}'_{i}$ is $\mathbf{A}_{i,b}$, $\mathbf{A}'_{i}$ has a form of a calibrated $2 \times 3$ camera matrix, and the problem with known gravity direction and vertical lines is equivalent to estimating a 2D structure from images taken by $n$ \textit{calibrated} cameras \cite{DBLP:journals/jmiv/AstromO00}.
Unlike its uncalibrated counterpart, the calibrated problem can be solved without projective ambiguity. Solvers for the cases $m=3, n=5$ and $m=4, n=4$ have been presented in \cite{DBLP:journals/jmiv/AstromO00}. Notably, these numbers exactly correspond to the minimal problems discussed in Sec.~\ref{sec:enumerating}.

The solver for $m=3, n=5$ estimates a trifocal tensor of the form \eqref{eq:tensor_elements} from $5$ point correspondences and $2$ linear internal constraints, and solves a quadratic function to decompose the tensor into camera matrices. The solver for $m=4, n=4$ estimates a dual quadrifocal tensor from $4$ point correspondences and $11$ linear internal constraints, and also solves a quadratic function to decompose the tensor into camera matrices. In both cases, there are two solutions in terms of matrices $\mathbf{A}'_{i}$. Typically, the correct solution is selected by counting the points lying in front of all cameras.

After finding $\mathbf{A}'_{i}$ with \cite{DBLP:journals/jmiv/AstromO00}, we construct the camera poses $\mathbf{R}_i(y_i), \mathbf{C}_i(y_i)$. For this, we first obtain $\mathbf{R}''_{i}$. Because \eqref{eq:matrix_constraint_calibrated} is homogeneous, there are two possible matrices $\mathbf{R}''_{i}$:
\begin{equation}
    \begin{split}
    \mathbf{R}''_{i} \in \left\{ \begin{bmatrix}
        (1-2 \mathbf{q}_{i,y}''^2) & 0 & 2\mathbf{q}_{i,w}''\mathbf{q}_{i,y}'' \\ 0 & 1 & 0 \\ -2\mathbf{q}_{i,w}''\mathbf{q}_{i,y}'' & 0 & (1-2 \mathbf{q}_{i,y}''^2)
    \end{bmatrix}, \right.\\
    \left.
    \begin{bmatrix}
       -(1-2 \mathbf{q}_{i,y}''^2) & 0 & -2\mathbf{q}_{i,w}''\mathbf{q}_{i,y}'' \\ 0 & 1 & 0 \\ 2\mathbf{q}_{i,w}''\mathbf{q}_{i,y}'' & 0 & -(1-2 \mathbf{q}_{i,y}''^2)
    \end{bmatrix} \right\} .
    \end{split}
     \label{eq:R_double_prime}
\end{equation}
The existence of two choices of $\mathbf{R}''_{i}$ per $\mathbf{A}'_{i}$ accounts for the presence of $16$ solutions in the $3$ view case and $32$ solutions in the $4$ view case.
%
%
Finally, we construct $\mathbf{R}_i(y_i), \mathbf{C}_i(y_i)$:
\begin{equation}
    \begin{split}
        \mathbf{R}_i(y_i) = \mathbf{R}_{0,i}^T \mathbf{R}_{a,i} \mathbf{R}''_{i},\\
        \mathbf{C}_i(y_i) = \begin{bmatrix}
        \mathbf{e}_1 & \mathbf{e}_3
    \end{bmatrix} \mathbf{A}_{i(1:2,1:2)}^{-1} \mathbf{A}_{i(1:2,3)}.
    \end{split}
     \label{eq:RC_final_E}
\end{equation}
Note, that $\mathbf{C}_i(y_i)_{(2)}$ is always zero. This is in accordance with the Section~\ref{sec:formulation}, which explains that the camera centers can only be estimated up to the shift along the direction $\mathbf{L}_d$.  

To summarize, we perform the following steps to solve problems $(E,3,5)$ and $(E,4,4)$:
\begin{enumerate}
    \item \textbf{Input}: Projections $\mathbf{p}_{i,j}$, vertical directions $\mathbf{v}_i$.
    \item Calculate matrices $\mathbf{R}_{0,i}, \mathbf{R}_{v,i}, \mathbf{R}_{a,i}$, and vectors $\mathbf{u}_{i,j}$ \eqref{eq:matrix_form_04}.
    \item Compose matrices $\mathbf{A}_{i,a}$ \eqref{eq:Aia} and vectors $\mathbf{u}_{i,j}'' \eqref{eq:matrix_constraint_calibrated}$.
    \item Use the $3$ view or the $4$ view solver from \cite{DBLP:journals/jmiv/AstromO00} to find $2$ sets of matrices $\mathbf{A}'_i$, $i \in \{1,...,m\}$ from $\mathbf{u}_{i,j}''$.
    \item Construct $\mathbf{R}_i''$ \eqref{eq:R_double_prime}, $\mathbf{R}_i(y_i)$, and $\mathbf{C}_i(y_i)$ \eqref{eq:RC_final_E}.
\end{enumerate}
On average, the $3$ view solver takes $9.16 \mu s$, and the $4$ view solver takes $70.84 \mu s$.

\subsection{Parallel lines and known gravity} \label{sec:parallel_lines_gravity}
We present the solution to the problem $(D,3,7)$. In this case, all lines share a common direction $\mathbf{L}_d$, which is generically not aligned with the vertical direction $\mathbf{e}_2$. Let $\mathbf{R}_d \in SO(3)$ be a rotation around an axis in the xz-plane, s.t. $\mathbf{L}_d = \mathbf{R}_d \mathbf{e}_2$. Substituting this into \eqref{eq:the_constraint}, we get
\begin{equation}
    \mathbf{p}_{i,j}^T \mathbf{R}_i(y_i) [\mathbf{R}_d \mathbf{e}_2]_{\times} (\mathbf{L}_{0,j} - \mathbf{C}_i(y_i)) = 0,
\end{equation}
\begin{equation}
    \mathbf{p}_{i,j}^T \mathbf{R}_i(y_i) \mathbf{R}_d [\mathbf{e}_2]_{\times} \mathbf{R}_d^T (\mathbf{L}_{0,j} - \mathbf{C}_i(y_i)) = 0. \label{eq:constraint_D_2}
\end{equation}
We define $\mathbf{L}_{0,j}' = \mathbf{R}_d^T \mathbf{L}_{0,j}$, $\mathbf{C}_{i}' = \mathbf{R}_d^T \mathbf{C}_i(y_i)$ and substitute these into \eqref{eq:constraint_D_2}:
\begin{equation}
    \mathbf{p}_{i,j}^T \mathbf{R}_i(y_i) \mathbf{R}_d [\mathbf{e}_2]_{\times} (\mathbf{L}'_{0,j} - \mathbf{C}'_i) = 0. 
\end{equation}
Following Sec.~\ref{sec:vertical_lines}, we find rotation matrices $\mathbf{R}_{0,i}$ and $\mathbf{R}_{v,i}$, decompose $\mathbf{R}_i(y_i)$ as $\mathbf{R}_{v,i} \mathbf{R}_{y,i}$, and find vectors $\mathbf{u}'_{i,j}$ according to \eqref{eq:matrix_form_05}. Then, we define $\mathbf{R}_{D,i} = \mathbf{R}_{0,i} \mathbf{R}_{v,i} \mathbf{R}_{y,i} \mathbf{R}_d$, and $\mathbf{t}_{D,i} = -\mathbf{R}_{D,i}^{-1} [\mathbf{e}_2]_{\times} \mathbf{C}'_i$, and rewrite the constraint as
\begin{equation}
    \lambda_{i,j} \mathbf{u}'_{i,j} = \begin{bmatrix} -\mathbf{R}_{D,i(1,3)} & \mathbf{R}_{D,i(1,1)} &  \mathbf{t}_{D,i(1)} \\ -\mathbf{R}_{D,i(3,3)} & \mathbf{R}_{D,i(3,1)} &  \mathbf{t}_{D,i(3)} \end{bmatrix} \begin{bmatrix}  \mathbf{L}'_{0,j(1)} \\ \mathbf{L}'_{0,j(3)} \\ 1 \end{bmatrix}, \label{eq:constraint_D_3}
\end{equation}
which can be written in matrix form as
\begin{equation}
    \lambda_{i,j} \mathbf{u}_{i,j} = \mathbf{A}_{D,i}\mathbf{L}'_{h,j}.
\end{equation}
This equation has the same form as \eqref{eq:matrix_form_05}, ensuring the existence of a trifocal tensor $\mathbf{T} \in \RR^{2,2,2}$ satisfying \eqref{eq:tensor_constraint}. We estimate this tensor linearly using $7$ correspondences, as in Sec.~\ref{sec:parallel_lines}.

Let us now discuss how to decompose $\mathbf{T}$ to estimate the relative pose. First, we fix the coordinate frame by setting $\mathbf{R}_1(y_1) = I$, $\mathbf{C}_1(y_1) = 0$.
Both the tensor $\mathbf{T}$ and the pose have $7$ DoF. Using Gr\"obner basis, we have discovered that for every tensor $\mathbf{T} \in \RR^{2,2,2}$, there are $48$ camera poses $\mathbf{R}_{y,2}, \mathbf{t}_{D,2}, \mathbf{R}_{y,3}, \mathbf{t}_{D,3}, \mathbf{R}_d$, such that the matrices $\mathbf{A}_{D,i}$ constructed via \eqref{eq:constraint_D_3} have the tensor $\mathbf{T}$.

Due to the high number of variables and equation complexity, we could not generate an ellimination template for a Gr\"obner basis solver. Therefore, we solve this problem with homotopy continuation \cite{DBLP:conf/cvpr/FabbriDFRPTWHGK20}, with an average runtime $19089 \mu s$.

Finally, we compose $\mathbf{A}_{i,D}$ according to \eqref{eq:constraint_D_3}, and the pose $\mathbf{R}_i(y_i), \mathbf{C}_i(y_i)$ as
\begin{equation}
\begin{split}
    \mathbf{R}_i(y_i) = \mathbf{R}_{v,i} \mathbf{R}_{y,i}, \\ \mathbf{C}_i(y_i) = R_d \begin{bmatrix}
        \mathbf{e}_1 & \mathbf{e}_3
    \end{bmatrix} \mathbf{A}_{D,i(1:2,1:2)}^{-1} \mathbf{A}_{D,i(1:2,3)}
\end{split}
     \label{eq:R_final_D}
\end{equation}

To summarize, we perform the following steps to solve the problem $(D,3,7)$:
\begin{enumerate}
    \item \textbf{Input}: Projections $\mathbf{p}_{i,j}$, vertical directions $\mathbf{v}_i$.
    \item Calculate matrices $\mathbf{R}_{0,i}, \mathbf{R}_{v,i}$ and vectors $\mathbf{u}_{i,j}$ \eqref{eq:matrix_form_04}.
    \item Construct linear system \eqref{eq:tensor_constraint} and solve it to get $\mathbf{T}$.
    \item Use homotopy continuation to decompose $\mathbf{T}$ into $\mathbf{R}_{y,2}$, $\mathbf{t}_{D,2}$, $\mathbf{R}_{y,3}$, $\mathbf{t}_{3,D}$, $\mathbf{R}_d$.
    \item Construct $\mathbf{R}_i(y_i)$ and $\mathbf{C}_i(y_i)$ using \eqref{eq:R_final_D}.
\end{enumerate}

\subsection{Degenerate scenarios}
Here, we analyze the degenerate cases for the solvers introduced in Sections \ref{sec:parallel_lines}, \ref{sec:vertical_lines}, and \ref{sec:parallel_lines_gravity}. Specifically, we consider scenarios where some lines are coplanar or where camera centers are collinear or coincident.
\begin{itemize}
    \item Solvers $(E,3,5)$, $(E,4,4)$ are degenerate if all $m$ lines are coplanar or if any two camera centers coincide.
    \item Solvers $(B,3,7)$, $(D,3,7)$ are degenerate if at least $5$ lines are coplanar or if any two camera centers coincide.
    \item There is no degeneracy if the camera centers are colinear.
\end{itemize}

\section{Experiments}

In this section, we evaluate the solvers $(B, 3, 7)$, $(D, 3, 7)$, $(E, 3, 5)$, $(E, 4, 4)$ on both synthetic and real data.

\subsection{Synthetic experiments}
\paragraph{Numerical stability.} We sample scanlines $y_i$ from $\mathcal{U}_{[-1,1]}$ and generate random camera rotations $\mathbf{R}_{i,GT}$ and centers $\mathbf{C}_{i,GT}$. 
Since we use one scanline per image, we only sample one pose per image.
We get the vertical directions as $v_i = \mathbf{R}_{i,GT} \mathbf{e}_2$.
To generate line $\mathbf{L}_j$, we sample scalars $\mu_{j,x}, \mu_{j,z}$ from normalized Gaussian distribution and set $\mathbf{L}_{0,j} = [\mu_{j,x} \ 0 \ \mu_{j,z}+5]^T$. For solver $(D,3,7)$, we also generate scalars $\nu_{x}, \nu_z$, and set $L_{j,d} = [\nu_{x} \ 1 \ \nu_{z}]^T$. In other cases, we use $L_{j,d} = \mathbf{e}_2$.
We generate $\mathbf{p}_{i,j}$ by projecting $\mathbf{L}_j$ into the camera with pose $(\mathbf{R}_{i,GT}, \mathbf{C}_{i,GT})$ and intersecting the projected line with scanline $y_i$.

Before measuring the error, we transform both the ground truth pose, and the estimated pose $(\mathbf{R}_{i,est}, \mathbf{C}_{i,est})$ into a canonical coordinate frame with $\mathbf{C}_1=0$, $\mathbf{R_{y,1}=\mathbf{I}, \mathbf{L}_d = \mathbf{e}_1}$. Since the y-coordinates of all camera centers are unobservable, we set them to zero. Then, we calculate the rotation error $err_\mathbf{R} = \max_{i=1}^m( \measuredangle \mathbf{R}_{i,est}^T \mathbf{R}_{i,GT} )$ and the translation error $err_\mathbf{C} = \max_{i=1}^m \measuredangle(\mathbf{C}_{i,GT}, \mathbf{C}_{i,est})$.

\begin{figure}
    \centering
    \input{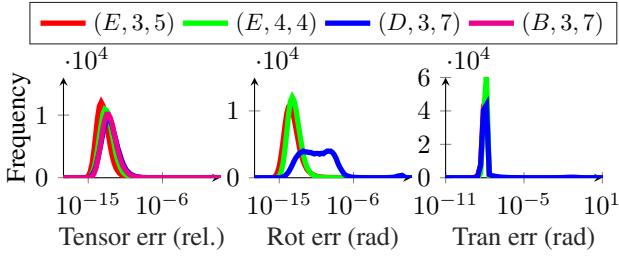}
    \vspace{-0.8cm}
    \caption{Histogram of tensor, rotation, and translation errors of the proposed solvers, over $10^5$ noiseless samples. In radians.}
    \label{fig:stability_tests}
\end{figure}

Additionally, we construct tensors $\mathbf{T}_{GT}$ from GT poses and estimate $\mathbf{T}_{est}$ using the solvers. For trifocal problems, $\mathbf{T}_{GT}, \mathbf{T}_{est}$ are trifocal tensors \eqref{eq:tensor_elements}, while for $(E,4,4)$, they are dual quadrifocal tensors. We measure tensor error $err_\mathbf{T}$ as the Frobenius norm of $\mathbf{T}_{GT} - \mathbf{T}_{est}$. For each solver, we generate $10^5$ synthetic noise-free scenes and measure $err_\mathbf{R}$, $err_\mathbf{C}$, and $err_\mathbf{T}$. Results in Fig.~\ref{fig:stability_tests} show solver stability.

\vspace{-0.2cm}

\begin{figure}[t]
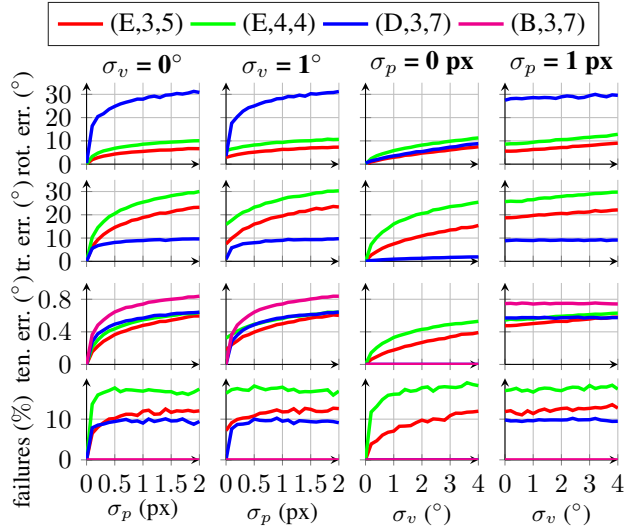

    \centering
    \setlength\tabcolsep{0pt}
    \renewcommand{\arraystretch}{0}
    \begin{tikzpicture}

\input{plot/R_P_0}
\input{plot/C_P_0}
\input{plot/T_P_0}
\input{plot/F_P_0}

\input{plot/R_P_D}
\input{plot/C_P_D}
\input{plot/T_P_D}
\input{plot/F_P_D}

\input{plot/R_V_0}
\input{plot/C_V_0}
\input{plot/T_V_0}
\input{plot/F_V_0}

\input{plot/R_V_D}
\input{plot/C_V_D}
\input{plot/T_V_D}
\input{plot/F_V_D}

\end{tikzpicture}
    \vspace{-0.7cm}
    \caption{
    Rotation, translation, and tensor errors averaged over 10000 successful runs as a function of the point and the vertical noise, measured from the real solution closest to the ground truth. If a solver returns only complex solutions, we report a failure. The failure rates are reported in the last row. Plots in the same row share the y-axis and plots in the same column share the x-axis. The fixed parameters are in the titles.
    }
    \label{fig:noise_tests}
\end{figure}

\paragraph{Tests with noise.} To evaluate the robustness to noise, we set focal length $f = 1000 px$, and add Gaussian noise with standard deviation $\frac{\sigma_p}{f}$ to each projection $\mathbf{p}_{i,j}$. We also perturb vertical directions $\mathbf{v}_i$ by applying a random 3D rotation with angle $\sigma_v$. Fig.~\ref{fig:noise_tests} shows $err_\mathbf{R}$, $err_\mathbf{C}$, and $err_\mathbf{T}$ as the function of $\sigma_p$, $\sigma_v$, averaged over $10^4$ runs. The results confirm that, since the line detections tend to be relatively precise, our solvers exhibit reasonable behaviour under realistic noise levels.

\subsection{Real-world Experiments}

We evaluate our solvers using the Fastec dataset \cite{Liu2020CVPR}, which consists of 19 sequences captured from a moving car in T\"ubingen. Each sequence contains 34 rolling shutter (RS) images and two global shutter (GS) images for every RS image: one with the pose of the first scanline and one with the pose of the middle scanline. The images have a resolution of $640 \times 480$ px and a focal length $768$ px.

Since GT poses are unavailable, we use COLMAP \cite{schoenberger2016sfm} on the GS images to obtain pseudo-GT poses $(\mathbf{R}_{i,F}, \mathbf{C}_{i,F})$ for the first scanline of image $i$ and $(\mathbf{R}_{i,M}, \mathbf{C}_{i,M})$ for the middle scanline. The pseudo-GT pose for scanline $y_i$ is interpolated as:
$ \mathbf{R}_{i,GT}(y_i) = \mathbf{R}_{i,M \rightarrow y_i} \mathbf{R}_{i,M},  
    \mathbf{C}_{i,GT}(y_i) = \mathbf{C}_{i,M} + \frac{h-2y_i}{h}(\mathbf{C}_{i,F}-\mathbf{C}_{i,M})
$, where $\mathbf{R}_{i,M \rightarrow y_i}$ is a rotation matrix with the axis of $\mathbf{R}_{i,F} \mathbf{R}_{i,M}^T$, and the angle $\frac{h-2y_i}{h}$ times the angle of $\mathbf{R}_{i,F} \mathbf{R}_{i,M}^T$.
We evaluate the solvers in \textit{multi-view} and \textit{single-view} settings.

\noindent \textbf{Multi-view.} We estimate poses for $m$ consecutive images, selecting one scanline per image. We use DeepLSD \cite{Pautrat_2023_DeepLSD} and Gluestick \cite{pautrat_suarez_2023_gluestick} to detect and match the lines.
Although the projections of lines by RS cameras are generically curves, the line-based methods work well in the case of Fastec dataset.
To select scanlines $y_1,...,y_m$, we exhaustively search for the scanlines that maximize the number of matched lines intersecting them.
Then, we find $p_{i,j}$ as the intersections of the detected lines and the scanlines.
Following \cite{StereoGlue2024}, we approximate the vertical direction by $[0 \ 1 \ 0 ]^T$. This setting works well for many dataset, including Fastec, and does not require gravity direction measurement.


\noindent \textbf{Single-view.} In this case, select $m$ scanlines in the same image and estimate the pose between them. We detect lines with DeepLSD \cite{Pautrat_2023_DeepLSD}, and select scanlines $y_1, ...,y_m$ with spacing at least $25$px with the highest number of shared lines. We obtain $p_{i,j}$ by intersecting lines with scanlines.
%


We evaluate the solvers in RANSAC \cite{ransac}. We label a correspondence as inlier, if the triangulated line $\mathbf{L}_j$ lies in front of all cameras and its reprojection error $\varepsilon$ is below $1px$. In the case of $(B,3,7)$, we decompose the tensor into canonical cameras \eqref{eq:canonical_cameras} before triangulation. Details about triangulation are in the SM. The score for a model is the sum over all lines of $(1px-\varepsilon)^2$ if $\varepsilon < 1px$, and zero otherwise.
We run RANSAC for $1000$ iterations, and select the model with the highest score. We do not use local optimization.

The results are summarized in Tables~\ref{tab:succ_rate_over_sequences},~\ref{tab:succ_rate_averaged} summarize the results for the solvers $(E,3,5)$, $(E,4,4)$, and $(D,3,7)$, and Table~\ref{tab:tensor_b37} shows the tensor error for the solver $(B,3,7)$.  
Although the total number of camera poses with an error below $10^\circ$ and $20^\circ$ remains relatively low, in Structure-from-Motion (SfM) applications, it is sufficient to estimate a single correct relative pose per sequence to initialize the reconstruction. Table~\ref{tab:succ_rate_over_sequences} confirms that this is achievable for the vast majority of sequences. A more detailed evaluation and further experiments are provided in the SM. 


\begin{table}[]
    \centering
    \begin{tabular}{c|c||c|c|c|c}
        \hline
        & solver & $\leq 5^{\circ}$ & $\leq 10^{\circ}$ & $\leq 20^{\circ}$ & $\leq 30^{\circ}$ \\
        \hline
        \multirow{3}{1.5em}{MV} & (E,3,5) & 1 & 3 & 10 & 15 \\
        & (E,4,4) & 5 & 8 & 11 & 14 \\
        & (D,3,7) & \textbf{6} & \textbf{10} & \textbf{15} & \textbf{17} \\
        \hline
        \multirow{3}{1.5em}{SV} & (E,3,5) & 3 & 3 & 5 & 10 \\
        & (E,4,4) & 2 & 4 & 7 & 13 \\
        & (D,3,7) & 4 & 4 & 7 & 9 \\
        \hline
    \end{tabular}
    \vspace{-0.3cm}
    \caption{Number of sequences (out of 19) from the Fastec dataset \cite{Liu2020CVPR}, with at least one estimated pose with error below $5/10/20/30^{\circ}$. MV: \textit{Multi-view}, SV: \textit{Single-view}. }
    \label{tab:succ_rate_over_sequences}
\end{table}

\begin{table}[]
    \centering
    \begin{tabular}{c|c||c|c|c}
        \hline
        & solver & $\leq 10^{\circ}$ & $\leq 20^{\circ}$ & $\leq 30^{\circ}$ \\
        \hline
        \multirow{3}{1.5em}{MV} & (E,3,5) & 1.0 \% & 6.2 \% & 21.9 \% \\
        & (E,4,4) & 1.9 \% & 7.9 \% & \textbf{22.4} \% \\
        & (D,3,7) & \textbf{5.6} \% & \textbf{13.8} \% & 21.7 \% \\
        \hline
        \multirow{3}{1.5em}{SV}
        & (E,3,5) & 0.5 \% & 2.5 \% & 5.3 \% \\
        & (E,4,4) & 1.2 \% & 4.1 \% & 10.0 \% \\
        & (D,3,7) & 0.5 \% & 2.2 \% & 6.6 \% \\
        \hline
    \end{tabular}
    \vspace{-0.3cm}
    \caption{Percentage of camera poses from Fastec \cite{Liu2020CVPR} with pose error below $10/20/30^{\circ}$. MV: \textit{Multi-view}, SV: \textit{Single-view}. }
    \label{tab:succ_rate_averaged}
\end{table}

\begin{table}[]
    \centering
    \begin{tabular}{c||c|c}
        \hline
        setting & Median error & Minimum error \\
        \hline
       MV & 0.85 & 0.44 \\
       SV & 0.89 & 0.62 \\
        \hline
    \end{tabular}
    \vspace{-0.3cm}
    \caption{Median and Minimum tensor error of the $(B,3,7)$ solver, averaged over all sequences from Fastec \cite{Liu2020CVPR}. A more detailed evaluation is in the SM.}
    \label{tab:tensor_b37}
\end{table}

\section{Limitations and Future Work}

While our approach shows promising results for rolling shutter relative pose estimation, several limitations remain. A key challenge is the lack of robust curve detectors and matchers, as rolling shutter distortion bends straight lines. Additionally, our method relies on chirality constraints to resolve ambiguity, as noted in \cite{DBLP:journals/jmiv/AstromO00}. The method does not use any motion model, which makes it broadly applicable and model-agnostic. However, this universality comes at the cost of stability: we expect methods that incorporate motion models to be more stable. Our solvers are also limited to specific scene types and assumes a known gravity direction. However, this reliance could be mitigated using scene knowledge, such as geometric relations between lines, in combination with solver $(B,3,7)$. Finally, our method estimates pose only up to a shift along the line direction. Future work could integrate additional geometric constraints, improve feature detection, and integrate the solvers to a rolling-shutter SfM pipeline.

\section{Conclusion}

In this paper, we propose a new approach for estimating the relative pose of rolling shutter cameras using projections of lines onto multiple scanlines. This enables estimating relative pose of rolling shutter cameras without explicitly modelling camera motion. The proposed solvers provide an important building block for rolling-shutter SfM where each scanline has its pose computed independently, without relying on a global motion model. 

We identify minimal problems for different variations of the single-scanline relative pose problem, including cases with parallel lines, vertical lines, and known gravity direction.
Furthermore, we develop minimal solvers for the parallel-lines scenario, both with and without gravity priors.
To evaluate our approach, we conduct experiments on both synthetic data and real-world rolling shutter images from the Fastec dataset, considering scanlines from either the same image or different images. These experiments demonstrate that our solvers successfully estimate at least one correct pose in the majority of scenes, confirming their feasibility for SfM initialization.
{
    \small
    \bibliographystyle{ieeenat_fullname}
    \bibliography{main}
}

\newpage
\appendix

\begin{center}
{\Large \textbf{Single-Scanline Relative Pose Estimation for Rolling Shutter Cameras}}\\
{\large Supplementary Material\\[2ex]
}
\vspace*{3ex}
\end{center}


\section{Decomposition of the Trifocal Tensor into Canonical Cameras}

Given a trifocal tensor $\mathbf{T} \in \mathbb{R}^{2,2,2}$, we describe that we can decompose it into two possible sets of $2 \times 3$ matrices $(\mathbf{A}_{c,1}, \mathbf{A}_{c,2}, \mathbf{A}_{c,3})$ in the canonical form. described in Eq.~(20) of the main paper.

First, we note that calculating the tensor elements according to Eq.~(23) of the main paper from cameras $\mathbf{A}_{c,1}, \mathbf{A}_{c,2}, \mathbf{A}_{c,3}$ yields
\begin{itemize}
    \item $T_{000} = 1$
    \item $T_{001} = \alpha_7$
    \item $T_{010} = \alpha_3$
    \item $T_{011} = \alpha_3 \alpha_7 - \alpha_4 \alpha_6$
    \item $T_{100} = \alpha_1$
    \item $T_{010} = \alpha_1 \alpha_7 - \alpha_1 \alpha_5$
    \item $T_{110} = \alpha_1 \alpha_3 - \alpha_1 \alpha_2$
    \item $T_{111} = \alpha_1 \alpha_3 \alpha_7 - \alpha_1 \alpha_4 \alpha_6 - \alpha_1 \alpha_2 \alpha_7 + \alpha_1 \alpha_4 \alpha_5 + \alpha_1 \alpha_2 \alpha_6 - \alpha_1 \alpha_3 \alpha_5 $
\end{itemize}
We normalize the tensor such that its first entry is $1$:
\begin{equation}
    \mathbf{T} \leftarrow \frac{\mathbf{T}}{T_{000}}.
\end{equation}
From the normalized tensor, we extract the parameters:
\begin{equation}
    \alpha_1 = T_{100}, \quad \alpha_3 = T_{010}, \quad \alpha_7 = T_{001},
\end{equation}
\begin{equation}
    \alpha_5 = \frac{-T_{101} + \alpha_1 \alpha_7}{\alpha_1}, \quad
    \alpha_2 = \frac{-T_{110} + \alpha_1 \alpha_3}{\alpha_1}.
\end{equation}
Now, $T_{100}$ and $T_{111}$ give two constraints for $\alpha_4$, $\alpha_6$. In the equation for $T_{111}$, we substitute $\alpha_3 \alpha_7 - T_{100}$ to get a linear constraint in $\alpha_4, \alpha_6$. Expressing $\alpha_6$ from this constraint and substituting into the constraint for $T_{100}$ yields a quadratic equation
\begin{equation}
    c_A \alpha_4^2 + c_B \alpha_4 + c_C = 0 \label{eq:SM_quadratic}
\end{equation}
with coefficients
\begin{equation}
\begin{split}
    c_A = -\frac{\alpha_5}{\alpha_2},\\
    c_B = \frac{\scriptstyle T_{111} - \alpha_1 \alpha_3 \alpha_7 + \alpha_1 (\alpha_3 \alpha_7 - T_{011}) + \alpha_1 \alpha_2 \alpha_7 + \alpha_1 \alpha_3 \alpha_5}{\alpha_1 \alpha_2},\\
    c_C = T_{011} - \alpha_3 \alpha_7.
\end{split}
\end{equation}
Solving \eqref{eq:SM_quadratic}, we obtain two solutions:
\begin{equation}
\begin{split}
    \alpha_{4a} = \frac{-c_B + \sqrt{c_B^2 - 4 c_A c_C}}{2 c_A}, \\
    \alpha_{4b} = \frac{-c_B - \sqrt{c_B^2 - 4 c_A c_C}}{2 c_A}.
\end{split}
\end{equation}
Finally, we find the corresponding values $\alpha_{6_a}$, $\alpha_{6_b}$ of $\alpha_6$ from the equation for $T_{111}$.

Using these values, we construct two sets of projection matrices:
\begin{equation}
    \mathbf{A}_1 =
    \begin{bmatrix}
        1 & 0 & 0 \\
        \alpha_1 & \alpha_1 & \alpha_1
    \end{bmatrix},
\end{equation}
\begin{equation}
    \mathbf{A}_2 =
    \begin{bmatrix}
        0 & 1 & 0 \\
        \alpha_2 & \alpha_3 & \alpha_4
    \end{bmatrix},
\end{equation}
\begin{equation}
    \mathbf{A}_3 =
    \begin{bmatrix}
        0 & 0 & 1 \\
        \alpha_5 & \alpha_6 & \alpha_7
    \end{bmatrix},
\end{equation}
where $(\alpha_4, \alpha_6)$ takes either $(\alpha_{4a}, \alpha_{6a})$ or $(\alpha_{4b}, \alpha_{6b})$, yielding two valid decompositions.

This method provides a direct way to extract the projection matrices from a given trifocal tensor while preserving the canonical structure. This also shows, that for a generic tensor $\mathbf{T} \in \RR^{2,2,2}$ there are two ways to decompose it into matrices in the canonical form.

\section{Triangulation of Lines from a Single Scanline}  

Given the projections of a 3D line onto a single scanline, we describe how we estimate a point $\mathbf{L}_0$ on the line and measure the reprojection error. Since the line direction $\mathbf{L}_d$ is known and corresponds to the y-axis, the problem reduces to estimating the 3D position of one point $\mathbf{L}_0$ on the line.  
Let $\mathbf{A}_i \in \mathbb{R}^{2 \times 3}$ be the projection matrix encoding the pose of the $i$-th camera, and let $\mathbf{u}_i \in \mathbb{R}^2$ be the measured projection of the line on the scanline. We construct the constraint matrix $\mathbf{M} \in \mathbb{R}^{m \times 3}$ as:  
\begin{equation}  
    \mathbf{M} =  
    \begin{bmatrix}  
        \frac{\mathbf{u}_1^T \mathbf{A}_1}{\|\mathbf{u}_1^T \mathbf{A}_1\|} \\  
        ... \\
        \frac{\mathbf{u}_m^T \mathbf{A}_m}{\|\mathbf{u}_m^T \mathbf{A}_m\|}  
    \end{bmatrix}.  
\end{equation}  

We solve for the homogeneous line representation $\mathbf{L}_h$ as the right singular vector corresponding to the smallest singular value of $\mathbf{M}$:  
\begin{equation}  
    \mathbf{L}_h = \arg \min_{\|\mathbf{L}\| = 1} \|\mathbf{M} \mathbf{L}\|.  
\end{equation}  
Normalizing $\mathbf{L}_h$ such that its third component is 1, we obtain:  
\begin{equation}  
    \mathbf{L}_h = \frac{\mathbf{L}_h}{\mathbf{L}_{h(3)}}.  
\end{equation}  
The estimated 3D point on the line is then given by:  
\begin{equation}  
    \mathbf{L}_0 =  
    \begin{bmatrix}  
        \mathbf{L}_{h(1)} \\  
        0 \\  
        \mathbf{L}_{h(2)}  
    \end{bmatrix}.  
\end{equation}  

To evaluate the reprojection error, we compute the projected coordinates of $\mathbf{L}_h$:  
\begin{equation}  
    \mathbf{p}_i = \mathbf{A}_i \mathbf{L}_h,  
\end{equation}  
and measure the angular deviation from the observed scanline projection:  
\begin{equation}  
    e_i = \left| \frac{p_{i,2}}{p_{i,1}} + \frac{u_{i,0}}{u_{i,1}} \right|.  
\end{equation}  
We get the total reprojection error by:  
\begin{equation}  
    e_{\text{total}} = \sum_{i=1}^m e_i^2.  
\end{equation}

\section{Additional experiments}

\subsection{Detailed evaluation of the experiments from the main paper}
Here, we present detailed results of the experiments conducted on the real-world dataset Fastec \cite{Liu2020CVPR}. In the case of solvers $(E,3,5)$, $(E,4,4)$, and $(D,3,7)$, we present median and minimum rotation and translation errors for every sequence from the Fastec dataset, as well as the percentages of scenes, whose pose error (obtained as the maximum over the rotation and translation error) does not exceed $10^{\circ}$, $20^{\circ}$, and $30^{\circ}$. In the case of solver $(B,3,7)$, we cannot measure the pose error, so we present the median and minimum tensor error for every sequence from the Fastec dataset.

\begin{table}[]
    \centering
    \setlength{\tabcolsep}{4pt}
    \begin{tabular}{c|c c | c c | c}
    \hline
    & \multicolumn{4}{c|}{R,t errors in $^\circ$} & Percentage of errors \\
    Seq & \multicolumn{2}{c |}{Median} & \multicolumn{2}{c |}{Minimum} & below 10$^\circ$ / 20$^\circ$ / 30$^\circ$  \\
    \hline
    4	&	1.0	&	71.8	&	6.0	&	12.0	&	0.0	/ 18.5 / 33.3 \\
    9	&	4.0	&	45.3	&	2.6	&	16.9	&	0.0	/	3.2	/	9.7 \\
    10	&	3.8	&	40.5	&	0.9	&	16.2	&	0.0	/	9.4	/	34.4 \\
    11	&	2.7	&	32.1	&	0.2	&	8.9	&	3.1	/	21.9	/	40.6 \\
    \hline
    13	&	3.2	&	38.1	&	0.9	&	7.5	&	3.1	/	12.5	/	12.5 \\
    16	&	4.1	&	42.4	&	3.1	&	32.7	&	0.0	/	0.0	/	0.0 \\
    17	&	2.0	&	38.6	&	1.0	&	13.8	&	0.0	/	12.5	/	31.3 \\
    18	&	6.0	&	44.7	&	2.2	&	20.3	&	0.0	/	0.0	/	25.0 \\
    20	&	2.4	&	42.8	&	1.0	&	20.0	&	0.0	/	0.0	/	31.3 \\
    \hline
    21	&	2.8	&	63.8	&	6.3	&	32.6	&	0.0	/	0.0	/	0.0 \\
    32	&	3.8	&	43.6	&	11.1	&	26.3	&	0.0	/	0.0	/	8.0 \\
    50	&	0.6	&	34.3	&	0.1	&	20.1	&	0.0	/	0.0	/	37.5 \\
    64	&	1.9	&	71.2	&	0.5	&	10.8	&	0.0	/	4.5	/	4.5 \\
    74	&	1.9	&	118.8	&	1.9	&	115.8	&	0.0	/	0.0	/	0.0 \\
    \hline
    76	&	0.8	&	111.4	&	0.3	&	51.3	&	0.0	/	0.0	/	0.0 \\
    79	&	1.4	&	25.2	&	0.4	&	10.7	&	0.0	/	15.6	/	62.5 \\
    88	&	3.4	&	40.0	&	1.8	&	27.1	&	0.0	/	0.0	/	10.0 \\
    92	&	2.2	&	25.9	&	0.6	&	3.3	&	12.5	/	12.5	/	59.4 \\
    100	&	4.0	&	44.3	&	7.6	&	19.0	&	0.0	/ 6.3 / 15.6 \\

    \hline
\end{tabular}
    \caption{Detailed evaluation of the errors for solver $(E,3,5)$ in the multi-view setting. For every sequence of Fastec \cite{Liu2020CVPR}, we give the median and the minimum rotation and translation error, as well as the percentage of pose errors below $10^\circ, 20^\circ, 30^\circ$.}
    \label{tab:table_A35}
\end{table}

\begin{table}[]
    \centering
    \setlength{\tabcolsep}{4pt}
    \begin{tabular}{c|c c | c c | c}
    \hline
    & \multicolumn{4}{c|}{R,t errors in $^\circ$} & Percentage of errors \\
    Seq & \multicolumn{2}{c |}{Median} & \multicolumn{2}{c |}{Minimum} & below 10$^\circ$ / 20$^\circ$ / 30$^\circ$  \\
    \hline
    
4	&	1.8	&	112.4	&	0.4	&	4.6	&	4.3	/	17.4	/	26.1	\\
9	&	5.3	&	43.8	&	1.1	&	10.6	&	0.0	/	10.0	/	16.7	\\
10	&	3.1	&	27.0	&	0.7	&	2.4	&	3.2	/	22.6	/	58.1	\\
11	&	3.8	&	36.1	&	3.8	&	7.8	&	3.2	/	22.6	/	35.5	\\
\hline
13	&	3.5	&	36.8	&	1.7	&	2.6	&	6.5	/	19.4	/	35.5	\\
16	&	6.9	&	50.5	&	3.7	&	32.3	&	0.0	/	0.0	/	0.0	\\
17	&	2.8	&	40.4	&	1.2	&	3.4	&	3.2	/	6.5	/	25.8	\\
18	&	7.0	&	71.3	&	4.5	&	36.5	&	0.0	/	0.0	/	0.0	\\
20	&	2.8	&	35.7	&	0.6	&	12.4	&	0.0	/	16.1	/	32.3	\\
\hline
21	&	5.2	&	70.4	&	3.0	&	9.4	&	3.2	/	3.2	/	3.2	\\
32	&	5.5	&	54.3	&	19.5	&	27.8	&	0.0	/	0.0	/	5.3	\\
50	&	2.2	&	62.4	&	1.1	&	17.9	&	0.0	/	6.5	/	22.6	\\
64	&	1.4	&	82.2	&	2.0	&	30.6	&	0.0	/	0.0	/	0.0	\\
74	&	5.1	&	115.5	&	5.1	&	115.5	&	0.0	/	0.0	/	0.0	\\
\hline
76	&	3.1	&	128.4	&	3.5	&	110.6	&	0.0	/	0.0	/	0.0	\\
79	&	2.3	&	43.0	&	0.7	&	21.7	&	0.0	/	0.0	/	20.7	\\
88	&	3.5	&	28.2	&	4.3	&	21.7	&	0.0	/	0.0	/	66.7	\\
92	&	4.6	&	26.4	&	2.3	&	4.0	&	9.7	/	22.6	/	54.8	\\
100	&	3.8	&	42.8	&	0.9	&	6.5	&	3.2	/	3.2	/	22.6	\\
    \hline
\end{tabular}
    \caption{Detailed evaluation of the errors for solver $(E,4,4)$ in the multi-view setting. For every sequence of Fastec \cite{Liu2020CVPR}, we give the median and the minimum rotation and translation error, as well as the percentage of pose errors below $10^\circ, 20^\circ, 30^\circ$.}
    \label{tab:table_A44}
\end{table}

\begin{table}[]
    \centering
    \setlength{\tabcolsep}{4pt}
    \begin{tabular}{c|c c | c c | c}
    \hline
    & \multicolumn{4}{c|}{R,t errors in $^\circ$} & Percentage of errors \\
    Seq & \multicolumn{2}{c |}{Median} & \multicolumn{2}{c |}{Minimum} & below 10$^\circ$ / 20$^\circ$ / 30$^\circ$  \\
    \hline
4	&	1.6	&	83.3	&	0.9	&	14.4	&	0.0	/	7.4	/	14.8	\\
9	&	8.6	&	55.0	&	16.8	&	24.6	&	0.0	/	0.0	/	3.2	\\
10	&	6.1	&	49.3	&	2.5	&	1.2	&	6.3	/	18.8	/	21.9	\\
11	&	10.4	&	59.7	&	2.1	&	7.3	&	6.5	/	22.6	/	25.8	\\
\hline
13	&	5.1	&	41.0	&	0.4	&	8.6	&	3.1	/	18.8	/	40.6	\\
16	&	13.0	&	57.7	&	6.2	&	35.4	&	0.0	/	0.0	/	0.0	\\
17	&	3.8	&	37.2	&	2.4	&	3.2	&	9.4	/	21.9	/	46.9	\\
18	&	8.1	&	93.5	&	3.0	&	24.2	&	0.0	/	0.0	/	16.7	\\
20	&	4.2	&	57.0	&	1.9	&	10.4	&	0.0	/	9.4	/	18.8	\\
\hline
21	&	6.3	&	85.8	&	4.2	&	4.8	&	3.1	/	9.4	/	12.5	\\
32	&	6.4	&	85.6	&	2.4	&	1.1	&	16.0	/	24.0	/	36.0	\\
50	&	3.2	&	94.7	&	1.5	&	12.5	&	0.0	/	3.1	/	9.4	\\
64	&	31.7	&	153.4	&	2.0	&	17.9	&	0.0	/	4.3	/	4.3	\\
74	&	180.0	&	180.0	&	180.0	&	180.0	&	0.0	/	0.0	/	0.0	\\
\hline
76	&	5.7	&	141.3	&	4.4	&	10.4	&	0.0	/	5.0	/	10.0	\\
79	&	6.1	&	44.8	&	2.1	&	2.8	&	3.1	/	15.6	/	28.1	\\
88	&	3.1	&	10.0	&	3.0	&	2.7	&	50.0	/	70.0	/	80.0	\\
92	&	6.8	&	53.8	&	6.5	&	9.7	&	6.3	/	25.0	/	28.1	\\
100	&	8.0	&	54.0	&	1.0	&	8.5	&	3.1	/	6.3	/	15.6	\\
    \hline
\end{tabular}
    \caption{Detailed evaluation of the errors for solver $(D,3,7)$ in the multi-view setting. For every sequence of Fastec \cite{Liu2020CVPR}, we give the median and the minimum rotation and translation error, as well as the percentage of pose errors below $10^\circ, 20^\circ, 30^\circ$.}
    \label{tab:table_D37}
\end{table}

\begin{table}[]
    \centering
    \setlength{\tabcolsep}{4pt}
    \begin{tabular}{c|c c | c c | c}
    \hline
    & \multicolumn{4}{c|}{R,t errors in $^\circ$} & Percentage of errors \\
    Seq & \multicolumn{2}{c |}{Median} & \multicolumn{2}{c |}{Minimum} & below 10$^\circ$ / 20$^\circ$ / 30$^\circ$  \\
    \hline
4	&	4.7	&	55.8	&	0.3	&	29.9	&	0.00	/	0.00	/	2.94	\\
9	&	0.5	&	117.8	&	0.5	&	2.6	&	3.03	/	3.03	/	6.06	\\
10	&	0.4	&	49.7	&	0.4	&	20.0	&	0.00	/	2.94	/	2.94	\\
11	&	1.3	&	52.2	&	3.0	&	12.0	&	0.00	/	5.88	/	5.88	\\
\hline
13	&	0.5	&	93.8	&	0.1	&	36.3	&	0.00	/	0.00	/	0.00	\\
16	&	7.1	&	129.6	&	0.2	&	28.1	&	0.00	/	0.00	/	4.00	\\
17	&	1.7	&	51.7	&	0.3	&	43.2	&	0.00	/	0.00	/	0.00	\\
18	&	89.9	&	128.7	&	5.2	&	41.8	&	0.00	/	0.00	/	0.00	\\
20	&	0.3	&	51.0	&	0.6	&	1.2	&	2.94	/	2.94	/	11.76	\\
\hline
21	&	0.3	&	29.2	&	0.2	&	4.3	&	2.94	/	32.35	/	52.94	\\
32	&	3.2	&	130.1	&	3.9	&	46.9	&	0.00	/	0.00	/	0.00	\\
50	&	4.4	&	125.9	&	1.2	&	47.7	&	0.00	/	0.00	/	0.00	\\
64	&	90.1	&	136.2	&	8.4	&	48.1	&	0.00	/	0.00	/	0.00	\\
74	&	179.9	&	179.9	&	179.9	&	179.9	&	0.00	/	0.00	/	0.00	\\
\hline
76	&	47.6	&	133.9	&	0.2	&	24.8	&	0.00	/	0.00	/	5.00	\\
79	&	0.7	&	115.5	&	8.1	&	23.9	&	0.00	/	0.00	/	2.94	\\
88	&	10.8	&	55.3	&	0.1	&	46.7	&	0.00	/	0.00	/	0.00	\\
92	&	0.7	&	133.3	&	0.3	&	37.6	&	0.00	/	0.00	/	0.00	\\
100	&	0.4	&	96.2	&	2.7	&	21.3	&	0.00	/	0.00	/	5.88	\\
    \hline
\end{tabular}
    \caption{Detailed evaluation of the errors for solver $(E,3,5)$ in the single-view setting. For every sequence of Fastec \cite{Liu2020CVPR}, we give the median and the minimum rotation and translation error, as well as the percentage of pose errors below $10^\circ, 20^\circ, 30^\circ$.}
    \label{tab:table_monocular_A35}
\end{table}

\begin{table}[]
    \centering
    \setlength{\tabcolsep}{4pt}
    \begin{tabular}{c|c c | c c | c}
    \hline
    & \multicolumn{4}{c|}{R,t errors in $^\circ$} & Percentage of errors \\
    Seq & \multicolumn{2}{c |}{Median} & \multicolumn{2}{c |}{Minimum} & below 10$^\circ$ / 20$^\circ$ / 30$^\circ$  \\
    \hline
4	&	1.4	&	81.2	&	1.7	&	1.5	&	11.11	/	18.52	/	22.22	\\
9	&	0.3	&	86.7	&	0.2	&	22.7	&	0.00	/	0.00	/	6.06	\\
10	&	0.3	&	49.8	&	1.2	&	6.0	&	3.03	/	9.09	/	15.15	\\
11	&	0.9	&	77.1	&	0.3	&	10.2	&	0.00	/	8.82	/	11.76	\\
\hline
13	&	0.6	&	111.6	&	0.0	&	51.5	&	0.00	/	0.00	/	0.00	\\
16	&	0.1	&	57.5	&	0.1	&	23.0	&	0.00	/	0.00	/	7.69	\\
17	&	0.7	&	71.0	&	0.5	&	3.0	&	5.88	/	14.71	/	17.65	\\
18	&	0.5	&	127.4	&	8.5	&	20.1	&	0.00	/	0.00	/	12.50	\\
20	&	0.3	&	59.5	&	0.7	&	11.6	&	0.00	/	11.76	/	17.65	\\
\hline
21	&	0.2	&	35.4	&	0.3	&	15.4	&	0.00	/	11.76	/	47.06	\\
32	&	0.3	&	49.8	&	1.7	&	22.2	&	0.00	/	0.00	/	10.53	\\
50	&	0.0	&	125.2	&	11.3	&	68.9	&	0.00	/	0.00	/	0.00	\\
64	&	1.3	&	134.6	&	3.7	&	27.0	&	0.00	/	0.00	/	7.69	\\
74	&	179.9	&	179.9	&	179.9	&	179.9	&	0.00	/	0.00	/	0.00	\\
\hline
76	&	0.3	&	74.0	&	1.0	&	26.8	&	0.00	/	0.00	/	11.11	\\
79	&	0.3	&	121.5	&	0.1	&	38.8	&	0.00	/	0.00	/	0.00	\\
88	&	0.0	&	47.5	&	0.0	&	46.6	&	0.00	/	0.00	/	0.00	\\
92	&	0.1	&	47.2	&	0.1	&	42.0	&	0.00	/	0.00	/	0.00	\\
100	&	0.2	&	87.3	&	0.4	&	5.2	&	2.94	/	2.94	/	2.94	\\
    \hline
\end{tabular}
    \caption{Detailed evaluation of the errors for solver $(E,4,4)$ in the single-view setting. For every sequence of Fastec \cite{Liu2020CVPR}, we give the median and the minimum rotation and translation error, as well as the percentage of pose errors below $10^\circ, 20^\circ, 30^\circ$.}
    \label{tab:table_monocular_E44}
\end{table}

\begin{table}[]
    \centering
    \setlength{\tabcolsep}{4pt}
    \begin{tabular}{c|c c | c c | c}
    \hline
    & \multicolumn{4}{c|}{R,t errors in $^\circ$} & Percentage of errors \\
    Seq & \multicolumn{2}{c |}{Median} & \multicolumn{2}{c |}{Minimum} & below 10$^\circ$ / 20$^\circ$ / 30$^\circ$  \\
    \hline
4	&	1.8	&	48.5	&	32.4	&	43.2	&	0.00	/	0.00	/	0.00	\\
9	&	0.2	&	52.9	&	1.5	&	3.9	&	3.03	/	6.06	/	15.15	\\
10	&	2.1	&	48.0	&	25.3	&	36.0	&	0.00	/	0.00	/	0.00	\\
11	&	7.6	&	48.4	&	1.1	&	17.6	&	0.00	/	2.94	/	5.88	\\
\hline
13	&	3.2	&	55.9	&	0.1	&	24.8	&	0.00	/	0.00	/	14.71	\\
16	&	2.7	&	50.1	&	14.6	&	43.4	&	0.00	/	0.00	/	0.00	\\
17	&	2.2	&	49.7	&	22.8	&	36.5	&	0.00	/	0.00	/	0.00	\\
18	&	6.9	&	56.8	&	6.6	&	26.2	&	0.00	/	0.00	/	4.17	\\
20	&	3.2	&	48.5	&	18.0	&	3.1	&	0.00	/	2.94	/	8.82	\\
\hline
21	&	1.5	&	28.4	&	2.1	&	3.0	&	2.94	/	20.59	/	61.76	\\
32	&	1.5	&	48.8	&	26.2	&	33.3	&	0.00	/	0.00	/	0.00	\\
50	&	10.3	&	60.5	&	4.4	&	49.9	&	0.00	/	0.00	/	0.00	\\
64	&	3.2	&	44.2	&	5.4	&	39.8	&	0.00	/	0.00	/	0.00	\\
74	&	180.0	&	180.0	&	180.0	&	180.0	&	0.00	/	0.00	/	0.00	\\
\hline
76	&	2.2	&	45.4	&	0.1	&	41.4	&	0.00	/	0.00	/	0.00	\\
79	&	0.4	&	63.5	&	1.2	&	19.5	&	0.00	/	2.94	/	8.82	\\
88	&	5.0	&	48.6	&	17.7	&	39.2	&	0.00	/	0.00	/	0.00	\\
92	&	1.3	&	46.1	&	22.8	&	34.0	&	0.00	/	0.00	/	0.00	\\
100	&	0.5	&	50.0	&	0.8	&	3.3	&	2.94	/	5.88	/	5.88	\\
    \hline
\end{tabular}
    \caption{Detailed evaluation of the errors for solver $(D,3,7)$ in the single-view setting. For every sequence of Fastec \cite{Liu2020CVPR}, we give the median and the minimum rotation and translation error, as well as the percentage of pose errors below $10^\circ, 20^\circ, 30^\circ$.}
    \label{tab:table_monocular_B37}
\end{table}

\begin{table}[]
    \centering
    \begin{tabular}{c||c|c}
        \hline
        Seq & Median tensor error & Minimum tensor error \\
        \hline
4	&	0.79	&	0.33	\\
9	&	0.74	&	0.23	\\
10	&	0.74	&	0.24	\\
11	&	0.63	&	0.20	\\
\hline
13	&	0.61	&	0.07	\\
16	&	0.77	&	0.63	\\
17	&	0.72	&	0.30	\\
18	&	0.93	&	0.46	\\
20	&	0.73	&	0.32	\\
\hline
21	&	1.12	&	0.72	\\
32	&	0.93	&	0.49	\\
50	&	0.67	&	0.40	\\
64	&	1.37	&	0.52	\\
74	&	1.41	&	1.41	\\
\hline
76	&	1.15	&	0.87	\\
79	&	0.52	&	0.10	\\
88	&	0.79	&	0.53	\\
92	&	0.57	&	0.25	\\
100	&	0.86	&	0.26	\\
        \hline
    \end{tabular}
    \caption{Detailed evaluation of the errors for solver $(B,3,7)$ in the multi-view setting. For every sequence of Fastec \cite{Liu2020CVPR}, we give the median and the minimum tensor error.}
    \label{tab:tensor_b37_big}
\end{table}

\begin{table}[]
    \centering
    \begin{tabular}{c||c|c}
        \hline
        Seq & Median error & Minimum error \\
        \hline
4	&	0.84	&	0.75	\\
9	&	0.87	&	0.06	\\
10	&	0.84	&	0.81	\\
11	&	0.81	&	0.32	\\
\hline
13	&	0.86	&	0.44	\\
16	&	0.86	&	0.83	\\
17	&	0.84	&	0.15	\\
18	&	0.90	&	0.53	\\
20	&	0.84	&	0.82	\\
\hline
21	&	0.47	&	0.08	\\
32	&	0.84	&	0.77	\\
50	&	0.91	&	0.88	\\
64	&	0.77	&	0.66	\\
74	&	2.00	&	2.00	\\
\hline
76	&	0.80	&	0.58	\\
79	&	1.05	&	0.40	\\
88	&	0.83	&	0.80	\\
92	&	0.80	&	0.78	\\
100	&	0.82	&	0.04	\\
        \hline
    \end{tabular}
    \caption{Detailed evaluation of the errors for solver $(B,3,7)$ in the single-view setting. For every sequence of Fastec \cite{Liu2020CVPR}, we give the median and the minimum tensor error.}
    \label{tab:tensor_b37_big_monocular}
\end{table}

\begin{figure}[]
    \centering
        \setlength{\tabcolsep}{4pt}
        \begin{tabular}{c c c c}
            \includegraphics[width=0.20\linewidth]{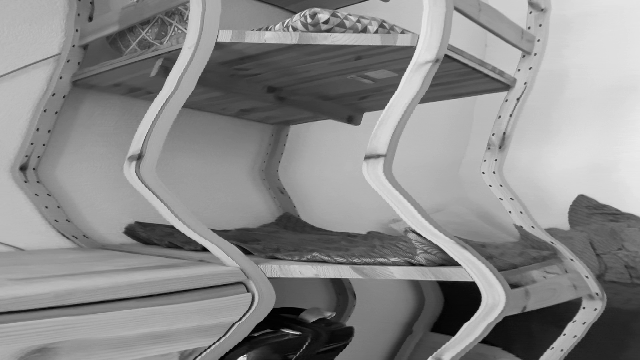} & \includegraphics[width=0.20\linewidth]{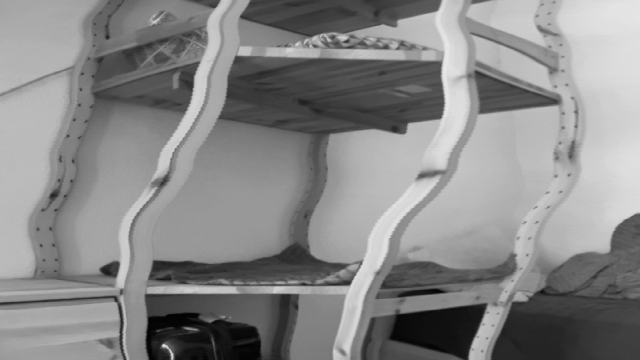} & \includegraphics[width=0.20\linewidth]{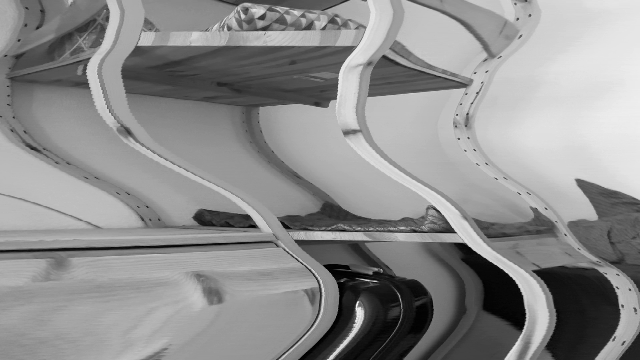} &
            \includegraphics[width=0.20\linewidth]{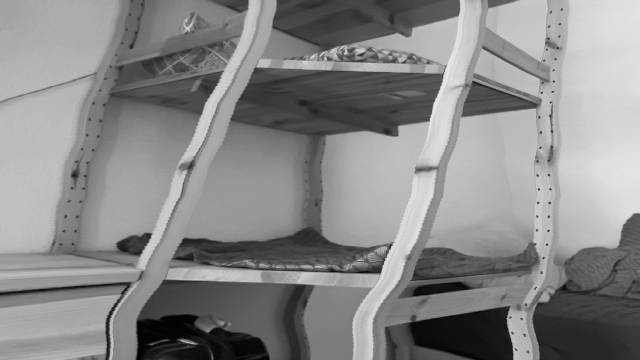} \\
            \includegraphics[width=0.20\linewidth]{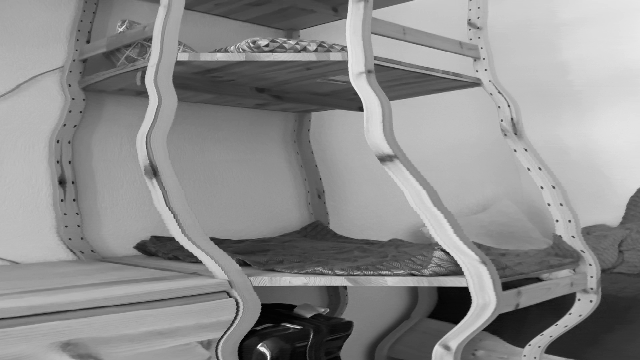} & \includegraphics[width=0.20\linewidth]{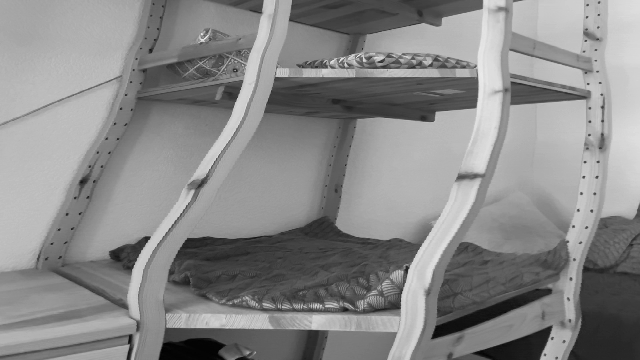} & \includegraphics[width=0.20\linewidth]{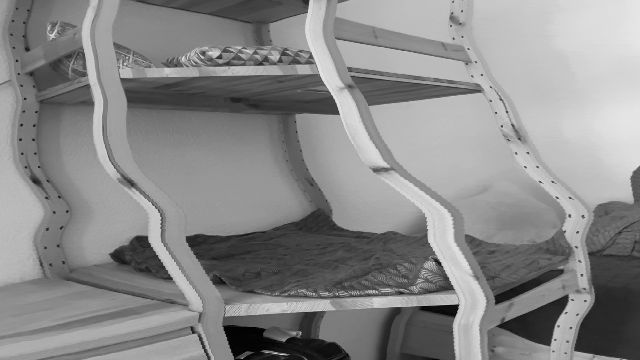} &
            \includegraphics[width=0.20\linewidth]{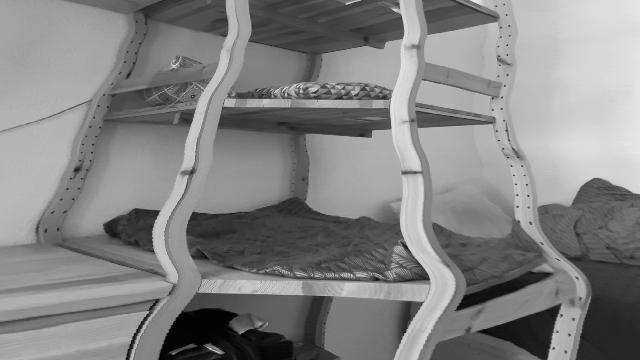} \\
        \end{tabular}
        \vspace{-0.4cm}
    \caption{Custom images used in the experiment in Section~\ref{sec:large_distortion}.}
    \label{fig:custom_images}
\end{figure}

\subsection{Evaluation on images with very large rolling-shutter distortions} \label{sec:large_distortion}

Here, we present an evaluation of our method on a custom dataset simulated very large rolling-shutter distortions. This scenario, while not very realistic, demonstrates that the proposed solvers are not limited to a particular type of motion.
We have taken multiple high speed videos of the same scene with an iPhone, and simulated rolling shutter images  (Fig.~\ref{fig:custom_images}) by selecting one scanline per frame. We have generated ground truth poses for every frame and obtained the camera intrinsics using COLMAP~\cite{schoenberger2016sfm}.

While it is very difficult to detect the whole lines, our method only needs to detect the segment of the line in the vicinity of a selected scanline. To detect these, we have found Canny edges in the images, and fitted conics into the edge points using \cite{halir1998numerically}. To match the curves, we obtained point matches using LoFTR~\cite{sun2021loftr}, associated every detected curve with keypoints within 2px from it, and used the number of shared associated keypoints as the score for matching the curves. 

We estimated the pose for every camera triplet from Fig.~\ref{fig:custom_images} using solver $(E,3,5)$. The median rotation and translation error are $4.8^{\circ}$ and $14.8^{\circ}$. $30.6 \%$ triplets have pose error below $10^{\circ}$, and $56.5\%$ have pose error below $20^{\circ}$. Every image is in at least one triplet with pose error below $6^{\circ}$.

We have further evaluated the five point solver \cite{DBLP:journals/pami/Nister04} on all images pairs from Fig.~\ref{fig:custom_images}. The median rotation and translation error are $6.6^{\circ}$ and $44.4^{\circ}$. $10.7 \%$ triplets have pose error below $10^{\circ}$, and $21.4\%$ have pose error below $20^{\circ}$.
This demonstrates that our method can handle large rolling-shutter distortions better than the five point solver.

\subsection{Pose estimation with local optimization}

\begin{figure}
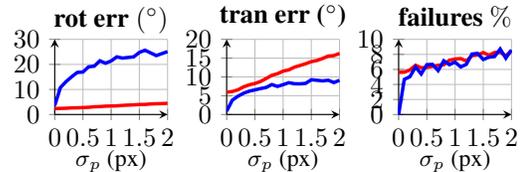

    \centering
    \begin{tikzpicture}

\input{plot/R_P_D_rebuttal}
\input{plot/C_P_D_rebuttal}
\input{plot/F_P_D_rebuttal}

\end{tikzpicture}
    \vspace{-0.45cm}
    \caption{Synthetic tests for solvers $(E,3,5)$ and $(D,3,7)$ with local optimization using $10$ lines. $\sigma_v = 1^{\circ}$.}
    
    \label{fig:synth}
\end{figure}

We have added local optimization (LO) after the solvers $(E,3,5)$ and $(D,3,7)$, and evaluated them on synthetic and real data. The result on synthetic data with $10$ lines per scene (Fig.~\ref{fig:synth}) shows that the LO improves the pose. However, the effect of the local optimization on real data from the Fastec dataset and from Fig.~\ref{fig:custom_images} is marginal, mostly due to a low number of lines in the images.

\end{document}